\documentclass[times,twocolumn,final]{elsarticle}

\usepackage{amsmath}
\DeclareMathOperator*{\argmax}{arg\,max}

\usepackage{medima}
\usepackage{framed,multirow}
\usepackage{mathtools}
\usepackage{amssymb}
\usepackage{latexsym}
\usepackage{graphicx}
\usepackage{comment}
\usepackage{amsmath,amssymb} 
\usepackage{color}
\usepackage{subfigure}
\usepackage[]{animate}
\usepackage{animate}
\usepackage{enumitem}
\usepackage{caption}
\usepackage{multirow}
\usepackage{amsthm}
\theoremstyle{plain}

\usepackage[section]{placeins}
\usepackage{soul}
\soulregister\cite7  
\soulregister\citep7  
\soulregister\citet7 
\soulregister\ref7 
\soulregister\pageref7  
\soulregister\cite7
\soulregister\citep7
\usepackage{multirow}
\usepackage{rotating}
\usepackage{wrapfig}
\usepackage{lineno}
\usepackage{epsfig}
\usepackage{nicefrac}       
\usepackage{microtype} 
\usepackage{float}
\usepackage{amsmath}
\usepackage{enumerate}
\usepackage{enumitem} 

\usepackage{url}
\usepackage{xcolor}

\usepackage{hyperref}

\definecolor{newcolor}{rgb}{.8,.349,.1}

\journal{Medical Image Analysis}

\begin{document}

\verso{X. Liu \textit{et~al.}}

\begin{frontmatter}

\title{Memory Consistent Unsupervised Off-the-Shelf Model Adaptation for Source-Relaxed Medical Image Segmentation}%


\author[1]{Xiaofeng  \snm{Liu} }

\author[1]{Fangxu \snm{Xing} }


\author[1]{Georges \snm{El Fakhri}}

\author[1]{Jonghye \snm{Woo}}

\address[1]{Gordon Center for Medical Imaging, Department of Radiology, Massachusetts General Hospital and Harvard Medical School, Boston, MA, 02114}

\received{20 Dec 2021,28 Apr 2022}

\accepted{16 Sep 2022}
\availableonline{}
\communicated{}

\begin{abstract}

Unsupervised domain adaptation (UDA) has been a vital protocol for migrating information learned from a labeled source domain to facilitate the implementation in an unlabeled heterogeneous target domain. Although UDA is typically jointly trained on data from both domains, accessing the labeled source domain data is often restricted, due to concerns over patient data privacy or intellectual property. To sidestep this, we propose ``off-the-shelf (OS)" UDA (OSUDA), aimed at image segmentation, by adapting an OS segmentor trained in a source domain to a target domain, in the absence of source domain data in adaptation. Toward this goal, we aim to develop a novel batch-wise normalization (BN) statistics adaptation framework. In particular, we gradually adapt the domain-specific low-order BN statistics, e.g., mean and variance, through an exponential momentum decay strategy, while explicitly enforcing the consistency of the domain shareable high-order BN statistics, e.g., scaling and shifting factors, via our optimization objective. We also adaptively quantify the channel-wise transferability to gauge the importance of each channel, via both low-order statistics divergence and a scaling factor.~Furthermore, we incorporate unsupervised self-entropy minimization into our framework to boost performance alongside a novel queued, memory-consistent self-training strategy to utilize the reliable pseudo label for stable and efficient unsupervised adaptation. We evaluated our OSUDA-based framework on both cross-modality and cross-subtype brain tumor segmentation and cardiac MR to CT segmentation tasks. Our experimental results showed that our memory consistent OSUDA performs better than existing source-relaxed UDA methods and yields similar performance to UDA methods with source data.

\end{abstract}


 \end{frontmatter}

\section{Introduction}

\begin{figure*}[t]
\begin{center}
\includegraphics[width=1\linewidth]{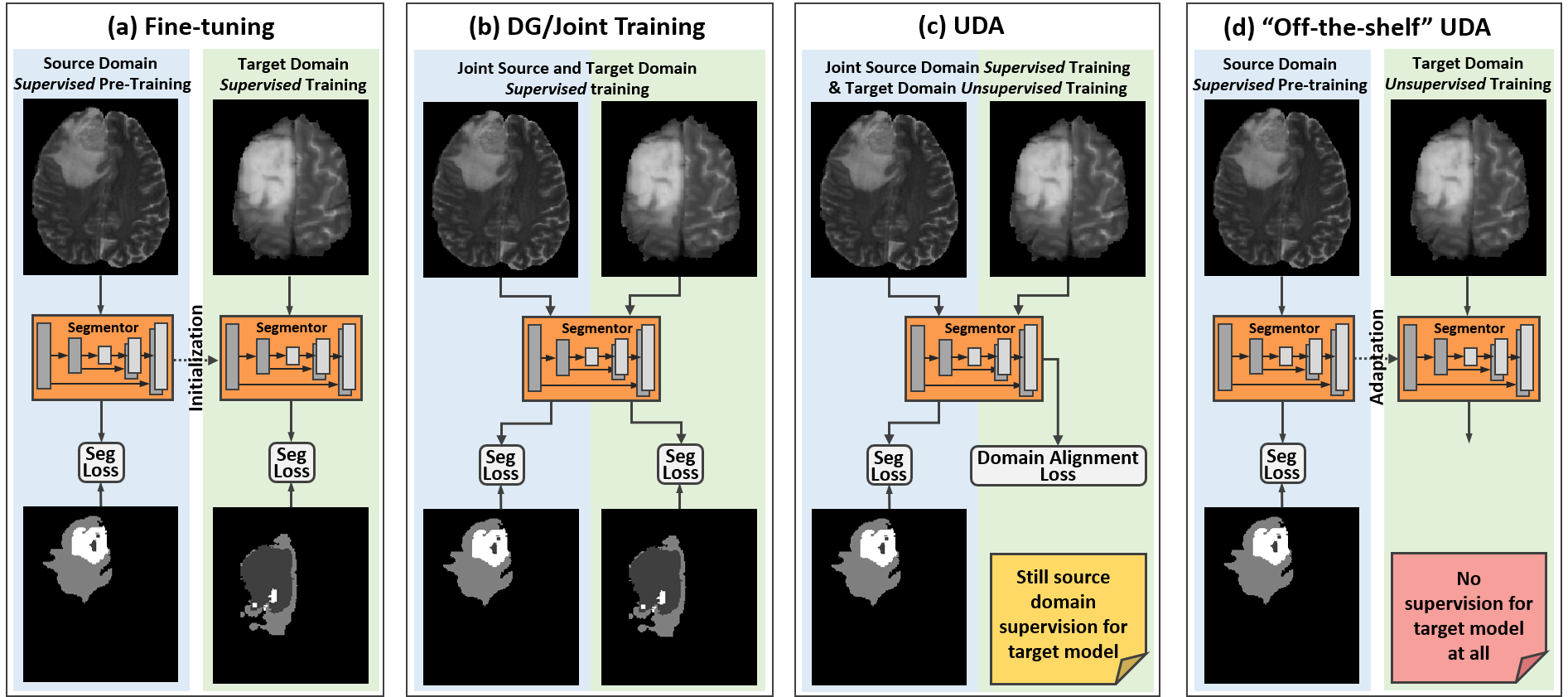}
\end{center}  \vspace{-10pt}
\caption{Comparisons of our framework with different related knowledge transfer methods: (a) fine-tuning makes use of labels in both domains via two stages, i.e., supervised pre-training in source domain and supervised re-training in target domain; (b) domain generalization (DG) \citep{liu2021Generalization} relies on joint training and expects generalization in unseen domains; (c) conventional UDA \citep{wilson2020survey,liu2021adversarial} is trained jointly on labeld source domain and unlabeled target domain data; and (d) our source-relaxed ``off-the-shelf" model adaptation for segmentation is based on adaptation, without source domain data.
}  
\label{fig1}\end{figure*} 

Accurate delineation of lesions or anatomical structures is a critical step for clinical intervention and treatment planning and has been markedly progressed over the past several years, mainly due to advances in deep neural networks (DNN) \citep{tajbakhsh2020embracing}. A deep segmentor trained on source domain data, however, cannot generalize well in a heterogeneous target domain, e.g., different clinical centers, subtypes, scanner vendors, or imaging modalities \citep{ghafoorian2017transfer,liu2021subtype}. Additionally, it often poses a great challenge to annotate labels for new target domain data \citep{che2019deep}. To mitigate these issues, unsupervised domain adaptation (UDA) has been proposed, as a promising technique, to achieve knowledge transfer from a labeled source domain to an unlabeled target domain \citep{liu2021generative,liu2021recursively}. 

Early attempts at UDA include statistic moment matching \citep{long2015fully}, feature/pixel-level adversarial learning \citep{liu2021adversarial}, and self-training \citep{zou2019confidence,liu2021generative}, all of which are dependent on the joint training on both source and target domain data. Well-labeled source domain data, however, are often inaccessible, due to concerns over patient data privacy or intellectual property (IP) \citep{bateson2020source}. As such, it is of great interest and needs to develop an adaptation strategy using an ``off-the-shelf (OS)" source domain model, without access to the source domain data. 

In recent years, a source-free UDA approach for classification \citep{liang2020we} was proposed to yield multiple predictions, but producing the diverse neighboring predictions is ill-suited for the segmentation purpose. A source relaxed UDA approach for segmentation \citep{bateson2020source} was also proposed to train an auxiliary class-ratio prediction model with source domain data, by relying on the assumption that the pixel proportion in segmentation is consistent between source and target domains. There are two major limitations in that work. First, the class-ratio in two domains can be different, due to label shift \citep{kouw2018introduction,liu2021adversarial,liu2021subtype,liu2021Generalization}, e.g., the incident rate of the disease and tumor size could vary depending on different tumor subtypes or populations. In addition, the class-ratio prediction model requires an extra training step with the source domain data. 

To address the aforementioned limitations, in this work, we propose a novel source-free UDA approach for segmentation, without an additional auxiliary network trained on the source domain data or reliance on the assumption of the same class proportion between two domains. Fig. \ref{fig1}~highlights the characteristics of conventional transfer learning approaches and our ``off-the-shelf" UDA (OSUDA). Our OSUDA does not need any label supervision in neither source nor target domain in adaptation, which works under a restrictive setting, compared with the conventional approaches \citep{long2015fully,liu2021adversarial,liu2021generative}.

~\\~
{In our prior work \citep{liu2021adapting}, we propose to leverage batch-wise normalization statistics that can be easily accessed and computed. Specifically, in modern deep learning backbones, such as ResNet \citep{He_2016_CVPR} and U-Net \citep{zhou2019normalization}, Batch Normalization (BN) \citep{ioffe2015batch} has been widely used to achieve fast and stable training. After training, the BN statistics are typically stored alongside the network parameters. Recent literature on source-available UDA methods has indicated that the low-order BN statistics, including the mean and variance in BN, are domain-specific, owing to the discrepancy of input data \citep{chang2019domain}. To achieve a gradual adaptation of the low-order BN statistics in a source-relaxed manner, a novel momentum-based progression strategy is presented, in which the momentum follows an exponential decay over the adaptation iteration. In addition, for a seamless transfer of the high-order BN statistics, including domain shareable scaling and shifting factors~\citep{maria2017autodial}, a high-order BN statistics consistency loss is proposed to enforce the discrepancy minimization. To this end, the transferability of each channel is measured first in an adaptive manner, followed by gauging the channel-wise importance. Further, unsupervised self-entropy (SE) minimization is used to enhance adaptation performance. 

In the present work, we build on our prior work \citep{liu2021adapting} as follows. First, we investigate and leverage the correlation between the high-order scaling factor and channel transferability for adaptive BN-based adaptation (in Sec. 3.2.3). Second, we develop a novel queued memory consistent OSUDA (MCOSUDA) in order to leverage the reliable pseudo label for stable and efficient adaptation (Secs. 3.4 and 3.5). Third, we further provide a detailed interpretation of our framework in the context of recent work in this area \citep{liu2021source,liu2021adversarial,kundu2021generalize,chen2021source,you2021domain}. Finally, in order to validate our framework, we experiment on brain tumor segmentation as well as cardiac segmentation tasks. New comparisons are also made with recent UDA work for all of the tasks to demonstrate the validity and superiority of our framework.}

Our contributions are summarized as follows:

\noindent$\bullet$ We propose a novel UDA segmentation framework in the absence of the source domain data, which thus only relies on an OS segmentor with BN trained on the source domain data. We do not need an additional auxiliary model trained in the source domain, or the assumption of the class-ratio consistency as in \cite{bateson2020source}.

\noindent$\bullet$ We systematically explore the domain-specific and shareable BN statistics, by means of the low-order BN statistics progression with an exponential momentum decay (EMD) strategy and the quantified transferability adaptive high-order BN statistics consistency loss, respectively.

\noindent$\bullet$ In addition to unsupervised SE minimization, we propose a queued memory consistent self-training for stable and efficient progression of the pseudo label in the target domain, via the memory-based supervision signal, conditioned on the historical consistency.

\noindent$\bullet$ We evaluate our framework on both (cross-modality and cross-subtype) brain tumor segmentation and cardiac MR to CT segmentation tasks using BraTS2018 and MM-WHS databases, to demonstrate the general efficacy of our framework and its superiority to conventional source-free/source-available UDA segmentation methods.

\section{Related Work}

\noindent\textbf{Unsupervised Domain Adaptation} has been widely used to migrate domain knowledge of one domain to another \citep{kouw2018introduction,liu2021adversarial}. Conventional approaches for UDA utilized the data in both domains for adaptation \citep{long2015fully,he2020classification,zou2019confidence}. This setting, however, requires the sharing of labeled source domain data, which poses a data privacy concern in real-world implementations. Recently, source-free UDA for classification \citep{li2020model,li2020free,liang2020we,wang2021tent} has been proposed to only use a pre-trained classifier, rather than co-training of the network with source and target domain data. Whereas these methods were applied only to classification or object detection, the present work focuses on segmentation tasks, i.e., fine-grained pixel-wise classification. CRUDA \citep{bateson2020source} pre-trained a class-ratio prediction model in the source domain, and enforced the pixel proportion consistency between two domains for segmentation. 

To the best of our knowledge, our prior work \citep{liu2021adapting} was the first attempt at OS source-relaxed UDA for segmentation, without the need for an additional auxiliary network, trained on the source domain data, or the assumption of the same class-ratio similar to CRUDA \citep{bateson2020source}. The follow-up works proposed methods based on knowledge distillation \citep{liu2021source} and source data hallucination \citep{ye2021source}, but an additional attention network or generative modules are required. In addition, \cite{kundu2021generalize} demanded a specific training strategy in the source domain, which does not utilize an OS segmentor, unlike our approach. Moreover, \cite{you2021domain} proposed a negative training strategy, designed specifically for classifying numerous classes, which is equivalent to the vanilla solution for a binary case. These works are also compared and discussed in relation to our proposed work. 

\vspace{+5pt}
\noindent\textbf{Batch Normalization} has been widely used to stabilize network training \citep{ioffe2015batch}, by eliminating an internal covariate shift. Early attempts of applying BN to adaptation simply added BN to a target domain, and did not have an interaction with the source domain \citep{li2018adaptive}. Recent work \citep{chang2019domain,maria2017autodial,wang2019transferable,mancini2018boosting} demonstrated that the low-order batch statistics, including the mean and variance in BN, are domain-specific, owing to the divergence of feature representations. Note that simply forcing the mean and variance in the target domain to be the same as the source domain can lead to a loss of expressiveness of the networks \citep{zhang2020generalizable}. In addition, once the low-order BN statistics discrepancy has been partially mitigated, the high-order BN statistics can be shareable between two domains \citep{maria2017autodial,wang2019transferable}. 

However, all of the aforementioned methods \citep{chang2019domain,maria2017autodial,zhang2020generalizable,wang2019transferable,mancini2018boosting} need joint training on source domain data. In the present work,  however, we rather opt to reduce the domain discrepancy, using a momentum-based adaptive low-order batch statistics progression strategy and explicit high-order BN statistics consistency loss for our source-free UDA for segmentation. 

\vspace{+5pt}
\noindent\textbf{Memory-based Learning} was initially proposed to stabilize supervised training with external modules to store memory \citep{weston2014memory}. Then, the idea of the memory mechanism has been generalized to semi-supervised learning \citep{tarvainen2017mean,laine2016temporal}, which utilized historical models as the regularization of the current network parameters. Therefore, we would expect more stable and competitive predictions. A moving-average model was used as the smoothed teacher model \citep{tarvainen2017mean} to regularize the UDA training \citep{french2017self,zheng2019unsupervised,luo2021unsupervised}. However, that work relied on the data in both domains for adaptation. Instead, in this work, we share a similar idea of the memory mechanism for efficiently stabilizing the OS adaptation.

\vspace{+5pt}
\noindent\textbf{Self-training} has been proposed to address semi-supervised learning \citep{triguero2015self}. Deep self-training was further proposed to integrate deep embedding learning and classifier adaptation \citep{zou2019confidence}. Recently, several deep self-training methods were proposed to utilize the pseudo label in the target domain data for UDA \citep{busto2018open,zou2019confidence,liu2020energy,liu2021generative}. However, the aforementioned works require co-training on source domain data, which is not applicable to our source-relaxed setting. Note that the conventional pseudo label generation can be highly unstable and unreliable, since it relies on the source domain supervision for correction \citep{zou2019confidence,liu2021energy}. Accordingly, in the present work, we used the self-training strategy with historical consistency-guided target data only learning.

\section{Methodology}

Let $\mathbf{x}\in\mathbb{R}^{H_0\times W_0\times C_0}$ be an input image with the height, width, and channel of $H_0$, $W_0$, and $C_0$, respectively. In segmentation, we predict the label of each pixel $\mathbf{y}_{h,w}\in\{1,2,\cdots, N\}$ as one of $N$ classes, yielding a segmentation map $\mathbf{y}\in\mathbb{R}^{H_0\times W_0}$. There are a source domain $\mathcal{D}^s$ and a target domain $\mathcal{D}^t$, where $\mathcal{D}^s\neq\mathcal{D}^t$ indicates that their inherited data distributions are different \citep{kouw2018introduction,liu2021adversarial}.
 
We assume that a segmentation model with BN, i.e., $f:\mathbf{x}\rightarrow\mathbf{y}$ parameterized with $\mathbf{w}$, is pre-trained with source domain samples $(\mathbf{x},\mathbf{y})\sim\mathcal{D}^s$. Note that the BN statistics are stored in the network itself, following the typical BN protocol. At the adaptation stage, we adapt the trained OS segmentor with only the unlabeled target domain samples $\mathbf{x}\sim\mathcal{D}^t$. 

 The domain adaptation theory \citep{ben2010theory} states that, for a hypothesis $h$ drawn from $\mathcal{H}$, the following condition is met: $\epsilon^t(h)\leq$ $\epsilon^s(h)+\frac{1}{2}d_{\mathcal{H}\triangle\mathcal{H}}\{s,t\}+e$, where, $\epsilon^s(h)$ and $\epsilon^t(h)$ represent the expected losses with hypothesis $h$ in source and target domains, respectively. The right side terms are considered an upper bound of the target loss. Of note, $e$ is usually a small and negligible value \citep{ben2007analysis}, and therefore UDA attempts to minimize the cross-domain divergence, $d_{\mathcal{H}\triangle\mathcal{H}}\{s,t\}$. Because the batch statistics are closely associated with the domain characteristics \citep{li2016revisiting,pan2018two}, we propose to leverage the transferability-aware batch-wise statistics for domain alignment, thereby offering a powerful inductive bias for target domain learning (in Subsec. 3.2). Specifically, in this work, the connection of the transferability between low and high-order statistics is investigated. In addition, the SE minimization of the target data prediction (in Subsec. 3.3) and the novel memory-consistence self-training (MCSF) strategy (in Subsec. 3.4) are further integrated as a unified framework (in Subsec. 3.5).

\subsection{Revisiting the Batch Normalization} 

BN has been widely used in modern deep learning models \citep{ioffe2015batch}. For a batch with $B$ images, which have the height, width, and channel in the $l$-th layer of $H_l, W_l,$ and $C_l$, the batch of input features in the $l$-th layer $f_l\in\mathbb{R}^{B\times H_l\times W_l\times C_l}$ is normalized for each channel. 

We index the samples in a batch with $b\in\{1,\cdots,B\}$, index the spatial position in the $l$-th layer with $m\in\{1,\cdots,H_l\times W_l\}$, and index the channel in the $l$-th layer with $c\in\{1,\cdots,C_l\}$, respectively. The channel-wise mean in BN is calculated by $\mu_{l,c}=\frac{1}{B\times H_l\times W_l}\sum_b^B\sum_m^{H_l\times W_l}f_{l,b,m,c}$, where $f_{l,b,m,c}\in\mathbb{R}$ is the feature value. Then, the channel-wise variance can be $\{\sigma^2\}_{l,c}=\frac{1}{B\times H_l\times W_l}\sum_b^B\sum_n^{H_l\times W_l} (f_{l,b,m,c}-\mu_{l,c})^2$. Note that the input feature is then normalized as $\overline{f}_{l,b,m,c}$, followed by applying a linear mapping:
\begin{align}
     \overline{f}_{l,b,m,c}&=\frac{{f}_{l,b,m,c}-\mu_{l,c}}{\sqrt{\{\sigma^2\}_{l,c}+\epsilon}}\\
     \tilde{f}_{l,b,m,c}&=\gamma_{l,c}\overline{f}_{l,b,m,c}+\beta_{l,c}, \label{bn}
\end{align}
where $\gamma_{l,c}$ and $\beta_{l,c}$ are learnable scaling and shifting factors. $\epsilon\in\mathbb{R}^+$ is a small scalar for numerical stability. After the normalization, $\overline{f}_{l,b,m,c}$ has zero mean and unit variance. In backpropagation, $\gamma_{l,c}$ and $\beta_{l,c}$ are updated with the gradient as the network parameters.

Instead of $B$ samples in a training batch, the testing input is usually a single sample. To bridge this gap, BN layer stores the weighted average of BN statistics in training and uses it for testing. Specifically, we use $k\in\{1,2,\cdots,K\}$ to index the training iteration, and the mean and variance in each iteration are tracked progressively following:
\begin{align}
\overline{\mu}_{l,c}^k&=(1-\eta)\cdot\overline{\mu}_{l,c}^{k-1}+\eta\cdot{\mu}_{l,c}^k,\\
\{\overline{\sigma}^2\}^k_{l,c}&=(1-\eta)\cdot\{\overline{\sigma}^2\}^{k-1}_{l,c}+\eta\cdot\{{\sigma}^2\}^k_{l,c},  
\end{align} where $\eta\in[0,1]$ is used to balance between the current and historical values. After $K$ training iteration, $\overline{\mu}_{l,c}^K$, $\{\overline{\sigma}^2\}^K_{l,c}$, $\gamma_{l,c}^K$, and $\beta_{l,c}^K$ are stored and used for the normalization in testing \citep{ioffe2015batch}.

\begin{figure}[t]
\begin{center}
\includegraphics[width=1\linewidth]{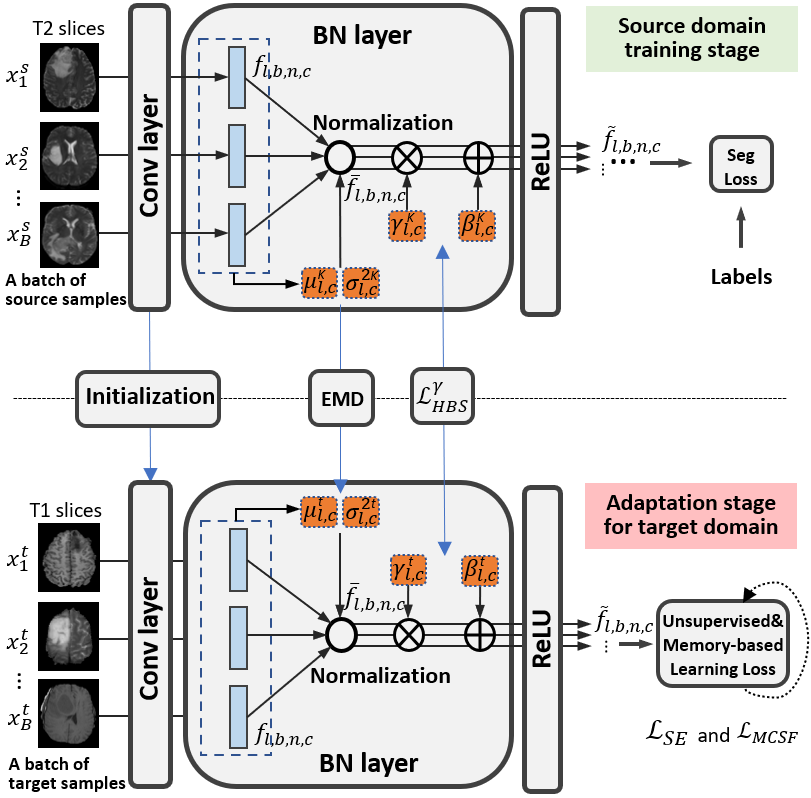}
\end{center}  
\caption{Illustration of a channel in our OSUDA framework, based on the pre-trained ``off-the-shelf (OS)" model with BN. We mitigate the domain discrepancy with the adaptive BN statistics in each channel.
} 
\label{fig2}\end{figure}

\subsection{Adaptive Source-relaxed Batch-wise Statistics Adaptation}

We propose to explore both the shared and domain-specific batch-normalization statistics with the quantified transferability to achieve the domain alignment \citep{liu2021adapting}. Specifically, we propose to reduce the domain divergence using an adaptive low-order BN statistics progression with EMD, and explicitly enforce the consistency of the high-order BN statistics in a source-free manner. In addition, the role of the scaling factor in transferability is further investigated.

\subsubsection{Low-order statistics progression with EMD}

For the target domain-specific factors, including mean and variance \citep{chang2019domain,maria2017autodial}, we propose a gradually learning scheme with EMD for low-order batch statistics progression in source-free UDA. First, the target domain mean and variance are initialized using the tracked $\overline{\mu}_{l,c}^K$ and $\{\overline{\sigma}^2\}^K_{l,c}$ in the source domain. In what follows, the target domain mean and variance in $t$-th adaptation iteration are progressively updated as
\begin{align}
\overline{\mu}_{l,c}^t&=(1-\eta^t)\cdot{\mu}_{l,c}^{t}+\eta^t\cdot\overline{\mu}_{l,c}^K, \\
\{\overline{\sigma}^2\}^t_{l,c}&=(1-\eta^t)\cdot\{{\sigma}^2\}^{t}_{l,c}+\eta^t\cdot\{\overline{\sigma}^2\}^K_{l,c},
\end{align} 
where $\eta^t=\eta^0\text{exp}(-t)$ is a momentum parameter with an exponential decay over iteration $t$. Note that ${\mu}_{l,c}^{t}$ and $\{{\sigma}^2\}^{t}_{l,c}$ are the mean and variance in the current target batch. Thus, the proportion of source statistics, i.e., $\overline{\mu}_{l,c}^K$ and $\{\overline{\sigma}^2\}^K_{l,c}$, are smoothly decreased along with the training, while ${\mu}_{l,c}^{t}$ and $\{{\sigma}^2\}^{t}_{l,c}$ gradually represent the low-order BN statistics in the target domain.    

\subsubsection{Transferability adaptive high-order statistics consistency}

For the domain shareable high-order BN statistics, including the learned scaling and shifting factors \citep{maria2017autodial,wang2019transferable}, we explicitly enforce their consistency across two domains in source-free UDA with a high-order BN statistics (HBS) loss, given by:
\begin{align}
\mathcal{L}_{HBS}= \sum_l^L\sum_{c}^{C_l} (1+\alpha_{l,c}) \{|\gamma_{l,c}^K-\gamma_{l,c}^t| + |\beta_{l,c}^K-\beta_{l,c}^t|\},
\end{align} where $\gamma_{l,c}^t$ and $\beta_{l,c}^t$ are the scaling and shifting factors in $t$-th adaptation iteration. We note that $\gamma_{l,c}^K$ and $\beta_{l,c}^K$are stored in the source domain model. $\alpha_{l,c}$ is used to adaptively balance among the channels.

The transferability of each channel can be different. For instance, \cite{pan2018two} demonstrated that channels with smaller low-order BN statistics divergence can be more transferable. As such, we propose a novel loss in a way that the channel with higher transferability contributes more to UDA. To this end, in order to quantify the  channel-wise domain discrepancy, the difference between batch statistics is measured as an efficient surrogate. For source-free UDA, we define a novel channel-wise cross-domain low-order BN statistics divergence at $t$-th adaptation iteration as
\begin{align}
    d_{l,c}=|\frac{\overline{\mu}_{l,c}^K}{\sqrt{\{\overline{\sigma}^2\}^K_{l,c}+\epsilon}}-\frac{{\mu}_{l,c}^t}{\sqrt{\{{\sigma}^2\}^t_{l,c}+\epsilon}}|.
\end{align} 
Then, the channel-wise low-order statistics divergence based transferability is quantified with $\alpha_{l,c}=\frac{L\times C\times(1+d_{l,c})^{-1}}{\sum_l\sum_c(1+d_{l,c})^{-1}}$. Therefore, the more transferable channels will have larger weight $(1+\alpha_{l,c})$ in $\mathcal{L}_{HBS}$, thereby contributing to the optimization with a higher importance. 


\subsubsection{Scaling factor for channel-wise transferability}

In addition to the mean and variance investigated in our prior work \citep{liu2021adapting}, we further explore the scaling factor $\gamma_{l,c}^K$ for quantifying the transferability. Conventionally, the scaling factor has been widely used as a criterion for the channel-wise importance in the channel pruning\footnote{\href{https://intellabs.github.io/distiller/tutorial-struct_pruning.html}{https://intellabs.github.io/distiller/tutorial-struct-pruning.html}} operations \citep{liu2017learning,kang2020operation}. The small scaling factor usually indicates the less effectiveness of this channel in a single domain classification task \citep{molchanov2019importance}. 

\begin{figure}[t]
\begin{center}
\includegraphics[width=1\linewidth]{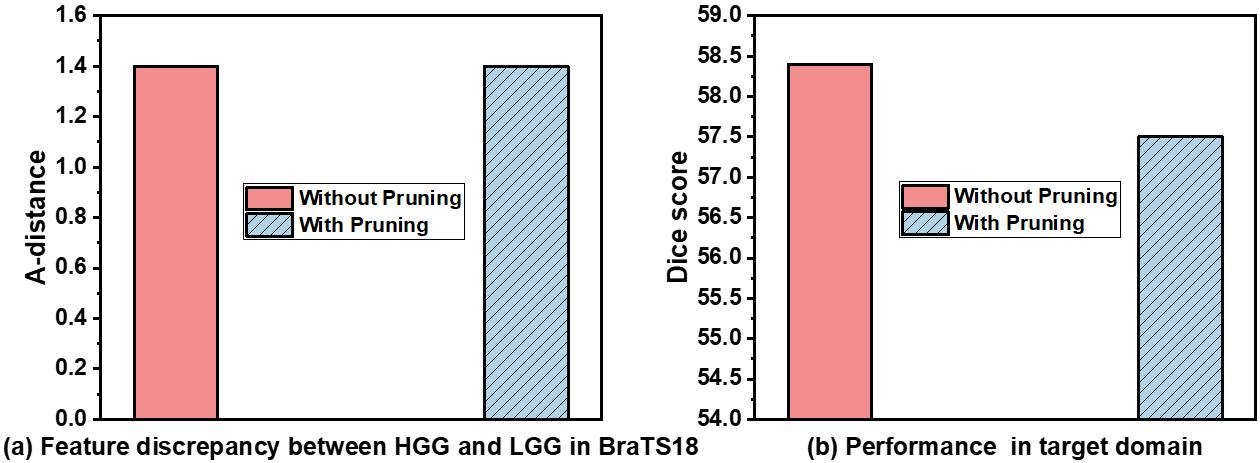}
\end{center} 
\caption{Comparison of (a) $\mathcal{A}-distance$ \citep{ben2010theory} and (b) the target domain DSC of the source domain model w/wo pruning of channels with a small scaling factor in the HGG to LGG task.} \vspace{-5pt}
\label{evid}
\end{figure}

Although these pruning works only focus on a single domain, we propose to investigate whether the transferability of each channel is also associated with its scaling factor. Specifically, we use the segmentation U-Net trained on high-grade gliomas (HGG) datasets and adapt it for low-grade gliomas (LGG) datasets in the BraTS2018 database \citep{menze2014multimodal} with our framework. $\mathcal{A}$-distance is a typical measurement for the feature discrepancy between two domains \citep{ben2010theory}, which has a smaller value for better alignment and 0 for  the same feature distribution. From Fig. \ref{evid}(a), we can see that the pruning of 10\% of the channels with the smallest $\gamma_{l,c}^K$ has the similar $\mathcal{A}$-distance as the network adapted without pruning, which indicates that these channels with small $\gamma_{l,c}^K$ may have little impact on UDA. This phenomenon can be simply explained from a gradient perspective. We denote all of the optimization objectives at the adaptation stage as $\mathcal{L}$. Then, the partial derivative of $\mathcal{L}$ w.r.t. $\overline{f}_{l,b,m,c}$ can be expressed as:
\begin{align}
    \frac{\partial\mathcal{L}}{\partial {\overline{f}_{l,b,m,c}}} = \frac{\partial\mathcal{L}}{\partial {\tilde{f}_{l,b,m,c}}}\gamma_{l,c}^K,
\end{align} 
which has a negligible gradient of ${\partial\mathcal{L}}/{\partial {\overline{f}_{l,b,m,c}}}$, if $\gamma_{l,c}^K$ is approaching to 0. Thus, the adaptation loss cannot efficiently enforce the domain alignment for this channel at the adaptation stage. 
 
We note that the channel dropping can be an efficient binary hard-weighting strategy to remove the less transferable channels, while it is difficult to define the reasonable threshold, thus rendering the network less expressive. Therefore, it can lead to a suboptimal solution. As shown in Fig. \ref{evid}(b), simply dropping 10\% of the channels will still lead to a performance drop in the target domain. Instead of simply dropping the channels with small $\gamma_{l,c}^K$, similar to prior work, we propose to leverage the scaling factor measured reliability via a soft-weighting strategy. Specifically, we utilize the scaling factor adjusted HBS loss:
\begin{align}
\mathcal{L}_{HBS}^\gamma= \sum_l^L\sum_{c}^{C_l} \exp{(-\gamma_{l,c}^K)}(1+\alpha_{l,c}) \{|\gamma_{l,c}^K-\gamma_{l,c}^t| + |\beta_{l,c}^K-\beta_{l,c}^t|\},
\end{align} 
where $\exp{(-\gamma_{l,c}^K)}$ has a smaller weight, if $\gamma_{l,c}^K$ is smaller. Thus, $\mathcal{L}_{HBS}^\gamma$ can take both low- and high-order BN statistics to achieve  channel-wise transferability quantification for adaptive BN-based adaptation.

\subsection{Self-entropy minimization in target domain} 
Although the label supervision in the target domain is not available, the unlabeled target domain can be guided by an unsupervised learning objective \citep{bateson2020source,liu2022Unsupervised}. 

A possible solution would be self-entropy (SE) minimization which has been a widely used objective in several deep learning models to enforce the confident prediction, i.e., the maximum softmax value can be high  \citep{grandvalet2005semi,liang2020we,wang2021tent,bateson2020source}. The pixel-wise SE for segmentation is formulated by the averaged entropy of the predictions of each pixel, given by
\begin{align}
    \mathcal{L}_{SE}=
    -\frac{1}{B\times H_0\times W_0}\sum_b^B\sum_m^{H_0\times W_0}\{{p_{b,m} \text{log} p_{b,m}}\},\label{se}
\end{align}
where $H_0$ and $W_0$ denote the height and width of the input, and $p_{b,m}\in\mathbb{R}^N$ indicates the $N$-class softmax output of the $m$-th pixel of the $b$-th image in a batch. Optimizing $\mathcal{L}_{SE}$ can encourage $p_{b,m}$ to approach to an one-hot vector. In addition, there are more alternative objectives to encourage confident predictions, e.g., the minimum class confusion (MCC) loss \citep{jin2020minimum}. We note that the MCC loss takes much more sophisticated computation than the classical SE loss, and is not scalable to the segmentation task \citep{jin2020minimum}. 

However, either SE loss or MCC loss only takes the prediction at the current iteration into consideration, which can be highly unreliable and unstable along with the training \citep{zou2019confidence}. In the case of source-free UDA, since there is no source data supervision in each iteration for correction, the biased prediction could significantly mislead the training.

\subsection{Queued memory-consistent Self-training} 

We further propose a novel queued memory-consistent self-training (MCSF) strategy for stable and efficient source-relaxed UDA. MCSF is able to stabilize the OSUDA from two perspectives. First, we only account for the pixel with high confident prediction in each iteration akin to conventional self-training \citep{zou2019confidence}. In addition, we calculate the supervision signal, conditioned on the historical consistency. 

Specifically, for a pixel in a target domain sample $\mathbf{x}_{b,m}$, we have the corresponding network prediction $p^{\tau}_{b,m}$ in $\tau$-th training iteration at the adaptation stage. The pseudo label $\hat{\mathbf{y}}_{b,m}=(\hat{{y}}^{1}_{b,m},...,\hat{{y}}^{N}_{b,m})$ of a target sample $\mathbf{x}_{b,m}$ is a $N$-dimensional vector indexed by $n$. The $n$-{th} dimension has the value of 1, only if the histogram distribution $p_{b,m}$ takes the maximum probability in $n$-th class, and the corresponding probability is larger than a class-wise threshold $\lambda^n$ \citep{zou2019confidence}. We note that $\lambda^n$ works as a confidence threshold to only keep the relatively reliable pseudo labels. We usually resort to the maximum value of $p_{b,m}$, as a surrogate of the confidence \citep{zou2019confidence}, and rank all of the pixels in a batch w.r.t. their maximum value of $p_{b,m}$. Then, $\lambda^n$ is set to select the top $\alpha\%$ of the most confident pixels in each class. Specifically, each class-wise bin of the pseudo label $\hat{\mathbf{y}}_{b,m}$ can be formulated as:
\begin{align}\label{psu}
\hat{{y}}^{n}_{b,m}=\left\{
\begin{aligned}
1, &~\text{if}~n=\argmax_{n}\{\frac {p_{b,m}(n|\mathbf{x},\mathbf{w}) }{\lambda^n}\} ~ \text{and} ~ p_{b,m}(n|\mathbf{x},\mathbf{w}) >\lambda^n,\\
0, &~\mathrm{otherwise},
\end{aligned}
\right.
\end{align}
where $p_{b,m}(n|\mathbf{x},\mathbf{w})$ is the value of $p_{b,m}$ in the $n$-th bin, which indicates the predicted probability of $n$-th class.~Therefore, $\hat{\mathbf{y}}^{n}_{b,m}$ can be a one-hot histogram for the reliable pixels, while $\hat{\mathbf{y}}^{n}_{b,m}$ can be 0 vector for the ones that are not confident. That is, the feasible set of $\hat{\mathbf{y}}_{b,m}\in \Delta^{N-1}\cup \{\mathbf{0}\}$ is a union of probability simplex $\Delta^{N-1}$ and $\{\mathbf{0}\}$. We note that the cross-entropy loss for $\hat{\mathbf{y}}_{b,m} = \{\mathbf{0}\}$ is always 0, which indicates that the corresponding pixel is not counted to the final loss.

\begin{figure}[t]
\begin{center}
\includegraphics[width=1\linewidth]{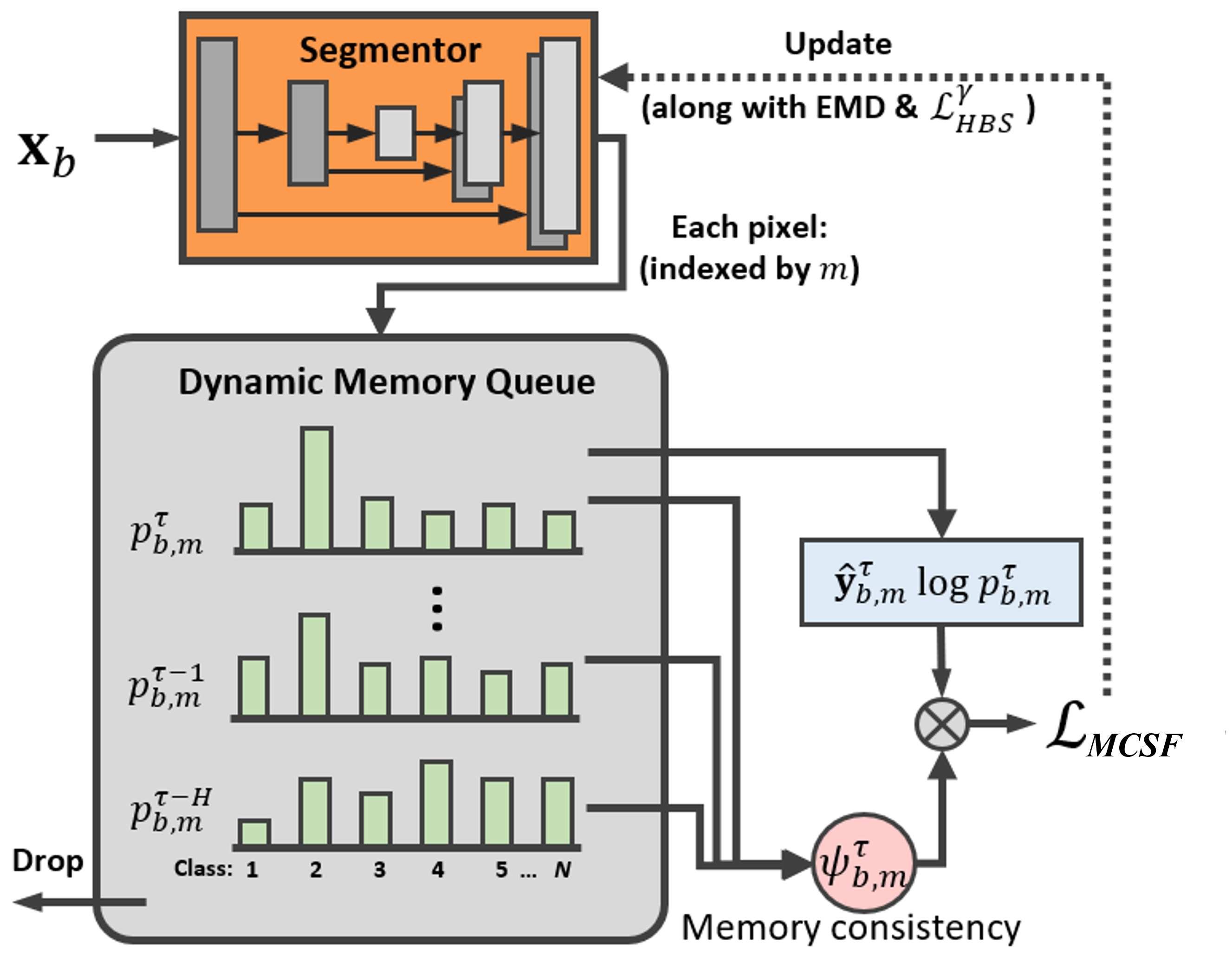}
\end{center} 
\caption{Illustration of our proposed queued dynamic memory-consistent self-training strategy.
} 
\label{memory}\end{figure}

\begin{table*}[t]
\centering
\caption{Comparisons of our framework against other UDA methods on the cross-MR-modality whole brain tumor UDA segmentation. T2-weighted MRI is used as the source domain, while T1-weighted, FLAIR, and T1ce MRI are used as the unlabeled target domains. The source-available UDA methods are regarded as ``upper bounds."}
\resizebox{0.9\linewidth}{!}{
\begin{tabular}{l|c|cccc|cccc}
\hline

&Source& & Dice & Score & [$\%$] $\uparrow$ &  & HD & [mm] $\downarrow$ \\ \cline{3-10}

{Method}&data& T1 & FLAIR & T1CE & Average & T1 & FLAIR & T1CE & Average \\ \hline \hline

Source only (no adatation)  & no UDA & 6.78 &54.36 &6.71& 22.62$\pm$0.176& 58.7& 21.5 &60.2 &46.8$\pm$0.15\\\hline

CRUDA~\citep{bateson2020source} & Partial & 47.25 &65.63& 49.47& 54.12$\pm$0.160 &22.1 &17.5 &24.4 &21.3$\pm$0.10\\

GTA \citep{kundu2021generalize} & Partial & 52.24 &64.03& 52.46& 56.24$\pm$0.135 &23.2 &16.7 &22.8 &20.9$\pm$0.11\\

DPL \citep{chen2021source} & \textbf{no} & 50.35 &63.17& 51.78& 55.10$\pm$0.121 &23.4 &17.0 &22.6 &21.0$\pm$0.09\\

SFKT \citep{liu2021source} & \textbf{no} & 51.71 &61.50&52.42& 55.31$\pm$0.197 &22.5 &16.9 &24.1 &21.2$\pm$0.07\\\hline  

\textbf{OSUDA}\citep{liu2021adapting} & \textbf{no} & {52.71} &{67.60}& {53.22}& {57.84$\pm$0.153}& {20.4} &{16.6}& {22.8}& {19.9$\pm$0.08}\\

OSUDA-AC & \textbf{no} & 51.58 &66.45 &52.12& 56.72$\pm$0.164 &21.5 &17.8 &23.6 &21.0$\pm$0.12\\    

OSUDA-SE & \textbf{no} & 51.14 &65.79& 52.80& 56.58$\pm$0.142 &21.6 &17.3 &23.3 &20.7$\pm$0.10\\ \hline

\textbf{OSUDA+}$\mathcal{L}_{HBS}^{\gamma}$ & \textbf{no} & {53.36} &{67.94}& {53.58}& {58.29$\pm$0.120}& {20.5} &{16.4}& {21.7}& {19.5$\pm$0.08}\\

\textbf{OSUDA+}$\mathcal{L}_{HBS}^{\gamma}$+MCSF (MCOSUDA)& \textbf{no} & \textbf{54.51} &\textbf{68.37}& \textbf{54.62}& \textbf{59.17$\pm$0.135}& \textbf{19.4} &\textbf{15.8}& \textbf{21.0}& \textbf{18.7$\pm$0.09}\\\hline \hline 

CycleGAN~\citep{zhu2017unpaired} & Yes&38.1 &63.3& 42.1& 47.8 &25.4 &17.2 &23.2 &21.9\\ 

SIFA~\citep{chen2019synergistic}& Yes& 51.7 &68.0 &58.2& 59.3 &19.6 &16.9& 15.01& 17.1\\

DSFN~\citep{zou2020unsupervised} & Yes& 57.3 &78.9 &62.2 &66.1& 17.5& 13.8 &15.5& 15.6\\ 

CLS~\citep{liu2021adversarial} & Yes& 56.92 &78.75 &63.24 &66.30$\pm$0.108& 17.2& 13.5 &15.8& 15.5$\pm$0.13\\ \hline
\end{tabular}}\label{tab2}
\end{table*}

Then, for each $\mathbf{x}_{b,m}^{\tau}$, we measure the pixel-wise historical consistency $\psi_{b,m}^{\tau}$ over $H$ consecutive iteration. Note that traversing all of the historical data can be inefficient to process. Therefore, we propose to incorporate the most recent $H$ iteration with a dynamic queue module that evolves smoothly alongside the network update. As shown in Fig. \ref{memory}, we utilize a ``first-in-first-out" memory queue to store $p_{b,m}^{\tau}$ and its previous values, i.e.,  $p_{b,m}^{\tau-1},\cdots,p_{b,m}^{\tau-H}$, to calculate the pixel-wise historical consistency $\psi_{b,m}^{\tau}$ at $\tau$-th iteration by:  
\begin{align} \label{mc}
\psi_{b,m}^{\tau} = 1 - \text{Sigmoid} (\frac{1}{H}\sum_{h=1}^H||p^{\tau}_{b,m}-p^{\tau-h}_{b,m}||_1).
\end{align}

We propose to utilize $\psi_{b,m}^{\tau}$ to achieve pixel-wise adaptive re-weighting of the self-training objective. In the case of a good agreement of the historical predictions, the target sample can be considered a well-learned one. Therefore, it is reasonable to rely on these samples and increase their contribution to the overall loss function, by assigning a large $\psi_{b,m}^{\tau}$. By contrast, we lower the weights for the largely historical inconsistent samples in the loss calculations. The training loss following the cross-entropy-based self-training can be formulated as:
\begin{align} 
\mathcal{L}_{MCST}= -\frac{1}{B\times H_0\times W_0}\sum_b^B\sum_m^{H_0\times W_0} \psi^{\tau}_{b,m}\times \hat{\mathbf{y}}^{\tau}_{b,m}\log p^{\tau}_{b,m},
\end{align} 
where $p^{\tau}_{b,m}$ is the prediction at $\tau$-th iteration, and $\hat{\mathbf{y}}^{\tau}_{b,m}$ is its pseudo label generated with Eq. (\ref{psu}).

Compared with the self-entrpy in Eq. (\ref{se}), $\mathcal{L}_{MCST}$ utilizes the pseudo label $\hat{\mathbf{y}}^{\tau}_{b,m}$ instead of $p^{\tau}_{b,m}$. Therefore, the low confident predictions will not be used to update the networks. In addition, the loss is adaptively adjusted with $\psi^{\tau}_{b,m}$.

\begin{figure}[t]
\begin{center} 
\includegraphics[width=1\linewidth]{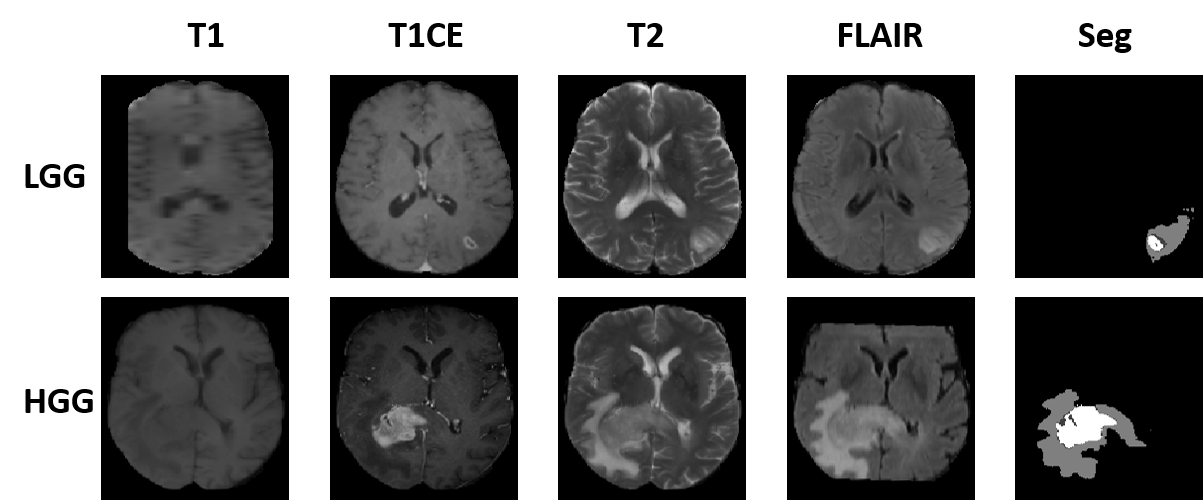}
\end{center}  
\caption{Illustration of the tumor size variability in the BraTS2018 database: the top row shows axial slices of LGG (top row) and HGG (bottom row) tumors with four MRI modalities and the corresponding segmentation label used in this work.}
\label{fige1}\end{figure}

\subsection{Overall training protocol for OSUDA with MCSF} 

At the target domain adaptation stage, the OSUDA with EMD and $\mathcal{L}_{HBS}^{\gamma}$ can be combined with either the conventional SE minimization or our novel queued memory-consistent self-training. Note that the training protocol of the source model, e.g., loss function and hyperparameters, is not a prerequisite for our model, but the trained model with its stored BN statistics alone is required. 
 
In the case of OSUDA with SE minimization, the overall training objective can be formulated as $\mathcal{L}=\mathcal{L}_{HBS}^{\gamma}+\lambda\mathcal{L}_{SE}$, where $\lambda$ balances between these two terms. However, SE may lead to a trivial solution that every sample has the same one-hot prediction \citep{grandvalet2005semi}. {For stabilization, we propose to change the contribution of the loss terms along with the training \citep{granger2020joint,ganin2016domain,tang2022unsupervised}. Specifically, we simply linearly decrease $\lambda$, e.g., from 10 to 0, along with the adaptation. Of note, the use of more sophisticated decreasing functions, e.g., the log or exponential function w.r.t. epoch \citep{granger2020joint,ganin2016domain} may further enhance adaptation performance.}

The optimization objective of the proposed OSUDA with MCSF can be expressed as 
\begin{align}\label{overal}
&\underset{\mathbf{w},{{{\hat{\mathbf{y}}}} }}{\mathop{\min }}\,  \mathcal{L}=\mathcal{L}_{HBS}^{\gamma}+\lambda\mathcal{L}_{SE}+\varphi\mathcal{L}_{MCSF}, \nonumber \\ 
&s.t.~{{{\hat{\mathbf{y}}}} }\in \Delta^{K-1}\cup \{\mathbf{0}\}, \forall~  \mathbf{x}\in\mathcal{D}^t, \text{~and~EMD},
\end{align}  
which is typically formulated as a classification maximum likelihood (CML) problem \citep{amini2002semi}, and can be optimized with Classification Expectation Maximization (CEM) \citep{zou2019confidence}. Specifically, there are three steps in each iteration \citep{zou2019confidence}:

\begin{figure*}[t]
\begin{center}
\includegraphics[width=1.03\linewidth]{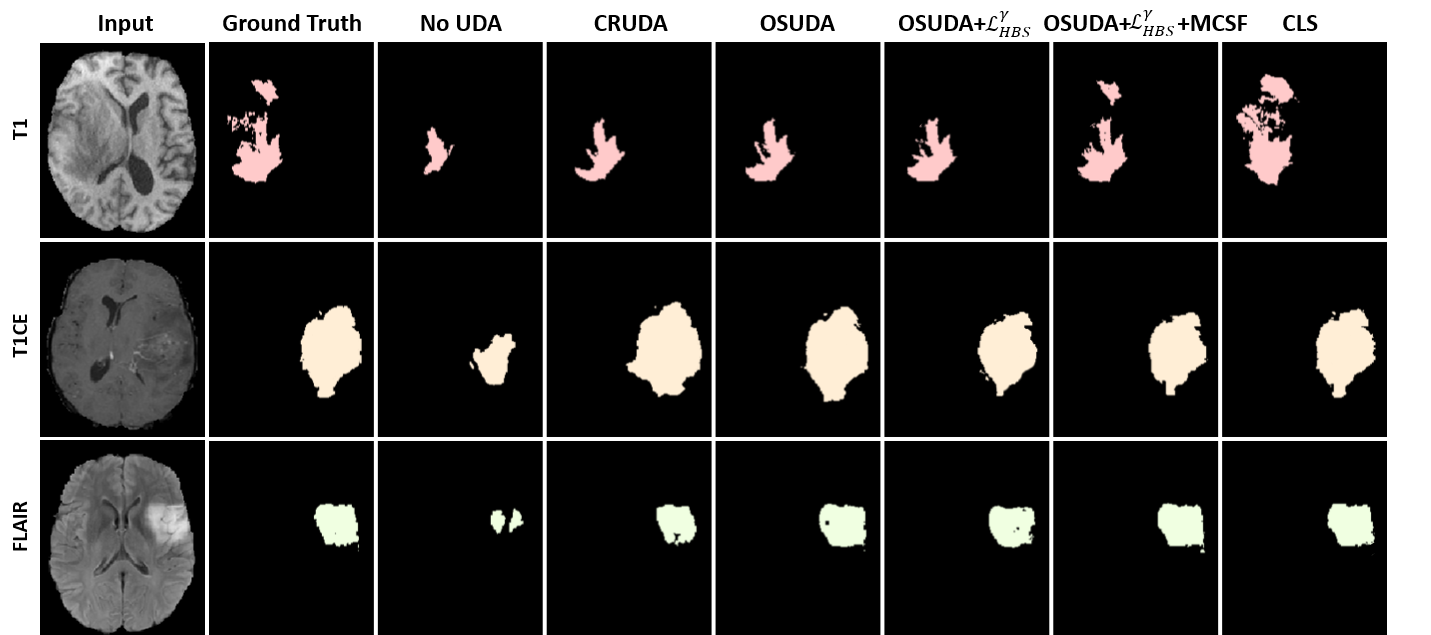}%
\end{center}  
\caption{Comparisons of our framework against other UDA methods, and ablation studies of T2-weighted MRI-to-T1-weighted/T1ce/FLAIR MRI UDA for whole tumor segmentation using BratS2018. Note that CLS \citep{liu2021adversarial} with source data for training is regarded as an ``upper bound."} 
\label{exp2}
\end{figure*}

\noindent\textbf{1. Expectation:} Estimating $p_{b,m}^{\tau}$ for all of $\mathbf{x}_{b,m}$ in a target batch with the forward pass of the current model.

\noindent\textbf{2. Classification:} Calculating the pseudo label $\hat{\mathbf{y}}^{\tau}_{b,m}$ (in Eq. (\ref{psu})) and the corresponding memory-consistency $\psi^{\tau}_{b,m}$ (in Eq. (\ref{mc})) for all of $\mathbf{x}_{b,m}$ in a target batch.

\noindent\textbf{3. Maximization:} Updating the low-order BN statistics with EMD, and fine-tuning the network parameter $\mathbf{w}$ with $\mathcal{L}_{HBS}^{\gamma}+\lambda\mathcal{L}_{SE}+\varphi\mathcal{L}_{MCSF}$ via backpropagation.

We note that solving for Eq.~(\ref{psu}) in our Expectation-classification steps is a typical concave problem that has a globally optimal solution. In addition, the Maximization step is seen as supervised learning, which is usually convergent \citep{shalev2014understanding,cover1999elements}. Thus, the overall training process can be convergent. In each iteration, we implement these three steps sequentially as the conventional self-training \citep{zou2019confidence,liu2021generative}.

\section{Experiments and Results}

To show the effectiveness of our framework, we experimented on both the brain tumor and cardiac segmentation tasks. We provided detailed comparisons against ``partial" source-relaxed UDA methods as well as the contemporary or follow-up source-free UDA methods. The source available UDA methods with the same backbones are used as our ``upper bounds." 

In addition, we provide ablation studies of the components in our framework, and the sensitivity analysis of hyperparameters invovled. We denote our prior work \citep{liu2021adapting} as OSUDA. The OSUDA without the adaptive channel-wise weighting and SE minimization are denoted as OSUDA-AC and OSUDA-SE, respectively. In addition, OSUDA+$\mathcal{L}^{\gamma}_{HBS}$ indicates using the scaling factor adjusted HBS loss for the high-order BN statistics consistency. We further integrate the memory-consistent self-training as our memory consistent OSUDA (MCOSUDA), i.e., OSUDA+$\mathcal{L}^{\gamma}_{HBS}$+MCSF, to achieve superior source-free adaptation performance.

We set $\epsilon=1\times10^{-6}$ for batch normalization as in \cite{ioffe2015batch}. Following the previous self-training works \citep{zou2019confidence,liu2020energy,liu2021generative}, we empirically initialize $\alpha=20$, and linearly increase it to 80 along with the training, since the pseudo-label is inherently more noisy at the start of training.

The training was performed with the PyTorch deep learning toolbox~\citep{paszke2017automatic} on an NVIDIA V100 GPU. For the evaluation metrics, we employed the widely accepted Dice similarity coefficient (DSC) and Hausdorff distance (HD) as in \cite{zou2020unsupervised,bateson2020source}. The DSC, a.k.a. dice score (the higher, the better), measures the overlap between the predicted segmentation mask and the label. The HD (the lower, the better) is defined for two sets of points in the prediction and label in a metric space \citep{zou2020unsupervised,bateson2020source}. The standard deviation was reported over five runs.

\subsection{Brain Tumor Segmentation}

The BraTS2018 database contains a total of 285 patients \citep{menze2014multimodal}, in which a total of 210 patients have glioblastoma, i.e., high-grade gliomas (HGG), while the remaining 75 patients have low-grade gliomas (LGG).

Each patient has 4 registered Magnetic Resonance (MR) Imaging (MRI) modalities, i.e., T1-weighted (T1), T1-contrast enhanced (T1ce), T2-weighted (T2), and T2 Fluid Attenuated Inversion Recovery (FLAIR) MRI. The MRI voxels have four class labels, i.e., enhancing tumor (EnhT), peritumoral edema (ED), tumor core (CoreT), and background. The union of CoreT, ED, and EnhT represents the whole tumor \citep{shanis2019intramodality}. To demonstrate the validity and generality of the proposed OSUDA, we carried out two cross-domain protocols, including cross-modality \citep{zou2020unsupervised} and cross-subtype UDA \citep{shanis2019intramodality}. {Fig.~\ref{fige1} shows example samples with the four MRI modalities from HGG or LGG datasets. Because there are imaging artifacts/low resolution in some slices (e.g., LGG T1 slice) and partly because of the cross-modality registration as a preprocessing \citep{menze2014multimodal}, some of the structures are incomplete. We note that the image volumes in BraTS have different resolution, and have been co-registered, interpolated to a standard resolution \citep{menze2014multimodal}.}

\begin{table*}[t!]
\caption{Comparisons of our framework against other UDA methods on cross-subtype, i.e., HGG to LGG UDA segmentation. The source-available methods, e.g., SEAT \citep{shanis2019intramodality}, are regarded as the ``upper bounds."} 
\label{tab1}
\centering
\resizebox{0.9\linewidth}{!}{
\begin{tabular}{l|c|cccc|cccc}
    \hline
    \multirow{2}*{Method}&Source& \multicolumn{4}{c|}{DSC [\%] $\uparrow$}& \multicolumn{4}{c}{HD [mm] $\downarrow$} \\ \cline{3-10}
    &data& WholeT & EnhT & CoreT & Overall & WholeT & EnhT & CoreT & Overall \\ \hline \hline
    Source only~\citep{shanis2019intramodality} & no UDA &79.29& 30.09& 44.11 &58.44$\pm$0.435 &38.7 &46.1 &40.2 &41.7$\pm$0.14\\\hline  
     
    CRUDA~\citep{bateson2020source} & Partial & 79.85 &31.05& 43.92& 58.51$\pm$0.123 &31.7 &29.5 &30.2 &30.6$\pm$0.15\\

    GTA \citep{kundu2021generalize} & Partial & 83.12 &31.92& 46.25& 61.38$\pm$0.152 &27.6 &24.8&27.4 &26.2$\pm$0.16\\

    DPL \citep{chen2021source} & \textbf{no} & 80.24 &31.26& 45.13& 59.72$\pm$0.136 &29.2 &26.5 &28.9 &28.0$\pm$0.13\\

    SFKT \citep{liu2021source}  & \textbf{no} & 81.79 &31.86& 46.42& 60.70$\pm$0.141 &29.0 &26.2 &28.1 &27.3$\pm$0.15\\    \hline

    \textbf{OSUDA}\citep{liu2021adapting} & \textbf{no} &{83.62} & {32.15}& {46.88} &{61.94$\pm$0.108} &{27.2} &{23.4} &{26.3} &{25.6$\pm$0.14}\\ 
    
    \textbf{OSUDA-AC} & \textbf{no} &82.74 & 32.04& 46.62 &60.75$\pm$0.145 &27.8 &25.5 &27.3 &26.5$\pm$0.16\\
    
    \textbf{OSUDA-SE} & \textbf{no} &82.45 & 31.95& 46.59 &60.78$\pm$0.120 &27.8 &25.3 &27.1 &26.4$\pm$0.14\\     \hline

     \textbf{OSUDA+}$\mathcal{L}_{HBS}^{\gamma}$ & \textbf{no} &{84.08} & {32.51}& {47.40} &{62.36$\pm$0.132} &{26.9} &{23.1} &{25.8} &{25.0$\pm$0.15}\\     
     
      \textbf{OSUDA+}$\mathcal{L}_{HBS}^{\gamma}$+MCSF (MCOSUDA) & \textbf{no} &\textbf{84.29} & \textbf{32.86}& \textbf{47.63} &\textbf{62.87$\pm$0.101} &\textbf{26.4} &\textbf{22.3} &\textbf{23.8} &\textbf{24.1$\pm$0.09}\\    \hline   \hline 
      
     SEAT~\citep{shanis2019intramodality} & Yes& 84.11 & 32.67 & 47.11 & 62.17$\pm$0.153 &26.4 &21.7 &23.5 &23.8$\pm$0.16\\ 
     
     CLS \citep{liu2021adversarial} & Yes&85.37 & 34.94& 49.25 & 64.02$\pm$0.135 &25.6 &21.0 &22.3 &22.9$\pm$0.15\\  \hline 
\end{tabular}
}
\end{table*}

\begin{figure*}[t]
\begin{center}
\includegraphics[width=1.02\linewidth]{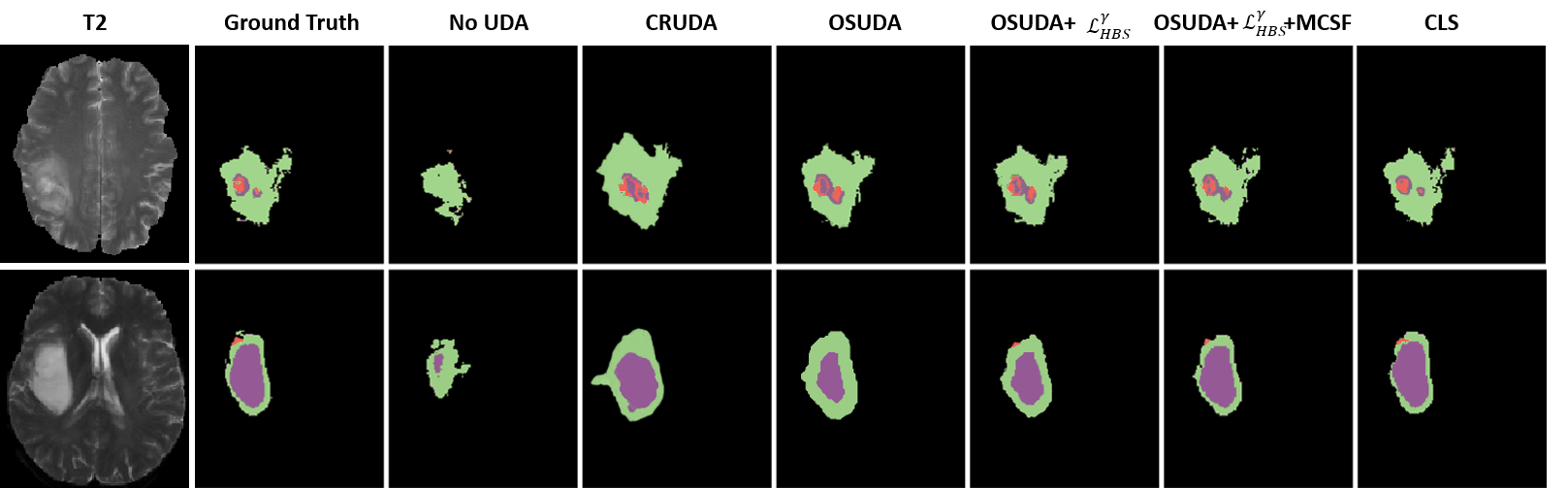}%
\end{center} 
\caption{Comparisons of our framework against other UDA methods, and ablation studies for HGG to LGG cross-subtype brain tumor UDA segmentation. The source-available CLS \citep{liu2021adversarial} is regarded as an ``upper bound."} 
\label{exp1}     
\end{figure*}

\subsubsection{Cross-MR-modality UDA: T2-weighted MRI to T1-weighted/T1ce/FLAIR MRI}

In the cross-modality UDA task, there can be relatively large appearance discrepancies among the MRI modalities. Considering the clinical manual annotation of the brain tumor usually works on T2-weighted MRI, the conventional cross-MR-modality UDA focuses on using T2-weighted MRI as the labeled source domain, while T1-weighted/T1ce/FLAIR MRI are used as the target domain \citep{zou2020unsupervised}. Following the standard protocol, we used 80\% subjects for training and 20\% subjects for testing \citep{zou2020unsupervised}, and used the same network backbone. All of the samples were used in a subject-independent and unpaired manner \citep{zou2020unsupervised}.

We followed the backbone as in \cite{zou2020unsupervised}. We set the batch size to 12 for segmentation. For all the adaptation models, a model trained on the source data with cross-entropy over 100 epochs was used as an initialization. The network was trained over 100 adaptation epochs. The consecutive iteration in the memory $H$ was set as 5. The training took about 1 hour. In practice, segmenting one image in testing only took about 0.1 seconds.

In Table~\ref{tab2}, we provide the quantitative evaluation results. The proposed OSUDA-based methods outperformed CRUDA \citep{bateson2020source} and the follow-up source-free UDA methods, e.g., GTA \citep{kundu2021generalize}, DPL \citep{chen2021source}, and SFKT \citep{liu2021source}. Note that CRUDA requires training an additional label proportion network in the source domain, and GTA requires training a special domain generalization module in the source domain, both of which are not source-free settings. In addition, our OSUDA-based methods with $\mathcal{L}_{HBS}^{\gamma}$ and MCSF outperformed the source-available UDA methods, e.g., CycleGAN \citep{zhu2017unpaired} and SIFA \citep{chen2019synergistic}, for T2-weighted MRI to T1-weighted/FLAIR MRI transfer tasks w.r.t. DSC and HD. We listed the state-of-the-art source available UDA methods as our ``upper bounds", and did not manage to beat all of them. Fig. \ref{exp2} illustrates the visual comparisons of 3 target MRI modalities, which shows the promising results of OSUDA-based methods over the other comparison methods. 

OSUDA-AC and OSUDA-SE yielded inferior performance than OSUDA, demonstrating the effectiveness of the adaptive channel-wise weighting and the SE minimization. In addition, the scaling factor adjusted $\mathcal{L}_{HBS}^{\gamma}$ and MCSF can further improve the performance.

\begin{table*}[t!]
\caption{Comparisons of our framework against other UDA methods on cardiac MR to CT segmentation. $\pm$ indicates standard deviation. The source available UDA methods are regarded as ``upper bounds."} 
\label{card1}
\centering
\resizebox{1\linewidth}{!}{
\begin{tabular}{l|c|ccccc|ccccc}
    \hline
    \multirow{2}*{Method}&Source& \multicolumn{5}{c|}{DSC [\%] $\uparrow$}& \multicolumn{5}{c}{HD [mm] $\downarrow$} \\ \cline{3-12}
    &data& AA & LAC & LVC & MYO & Average & AA & LAC & LVC & MYO & Average \\ \hline \hline
    
    Source only~\citep{zou2020unsupervised} & no UDA &28.4 & 27.7 & 4.0 & 8.7 & 17.2 & 32.6  &35.1  &54.2 & 57.4  &44.8\\\hline  
     
    CRUDA~\citep{bateson2020source} & Partial & 74.94 &72.33& 58.26& 30.40&58.98$\pm$0.109 &21.7 &22.1 &24.3 &30.7&24.7$\pm$0.14\\
    
    SFKT \citep{liu2021source}  & \textbf{no} & 75.83 &71.49& 59.76&34.58 &60.41$\pm$0.152 &19.4 &18.5 &23.2&29.1 &22.5$\pm$0.11\\    \hline  
    
    \textbf{OSUDA}\citep{liu2021adapting} & \textbf{no} &{78.02} & {73.55}& {60.18}& 40.68&{63.11$\pm$0.160} &{11.6} &{15.1} &{19.4}& 20.7&{16.7$\pm$0.13}\\ 
    
    \textbf{OSUDA-AC} & \textbf{no} &72.89 &73.43& 60.02& 40.41&62.94$\pm$0.145 &12.0 &16.4 &19.9& 21.3&17.4$\pm$0.16\\
    
    \textbf{OSUDA-SE} & \textbf{no} &77.85 & 73.36& 60.10& 40.17&62.87$\pm$0.142 &12.8 &16.5 &19.4& 21.7&17.6$\pm$0.15\\     \hline      
           
    \textbf{OSUDA+}$\mathcal{L}_{HBS}^{\gamma}$ & \textbf{no} &{78.16} & {73.82}& {60.39}& 41.46&{63.46$\pm$0.117} &{11.4} &{14.2} &{18.4}& 20.1&{16.0$\pm$0.12}\\     
    \textbf{OSUDA+}$\mathcal{L}_{HBS}^{\gamma}$+MCSF (MCOSUDA)& \textbf{no} &\textbf{78.80} & \textbf{75.63}& \textbf{61.45}&\textbf{42.38} &\textbf{64.57$\pm$0.195} &\textbf{10.7} &\textbf{11.9} &\textbf{18.3}& \textbf{19.5}&\textbf{15.1$\pm$0.10}\\    \hline   \hline 
      
CycleGAN~\citep{zhu2017unpaired} & Yes&73.8 &75.7 &52.3 &28.7 &57.6& 16.2& 15.4& 20.8 &27.7& 20.0\\ 

SIFA~\citep{chen2019synergistic}& Yes& 81.1 &76.4 &75.7 &58.7 &73.0 &8.2 &10.5 &12.2 &17.9& 12.2\\

DSFN~\citep{zou2020unsupervised} & Yes& 84.7 & 76.9  &79.1  &62.4 & 75.8 & 7.4 & 11.9 & 10.6 & 15.7 & 11.4\\ 

CLS~\citep{liu2021adversarial} & Yes& 84.82 &77.30  &79.04  &62.65 & 75.95$\pm$0.138 & 7.6 & 11.7 & 10.4 & 15.5 & 11.3$\pm$0.15\\ 
\hline 

\end{tabular}
}
\end{table*}

\begin{figure*}[t]
\begin{center}
\includegraphics[width=1.03\linewidth]{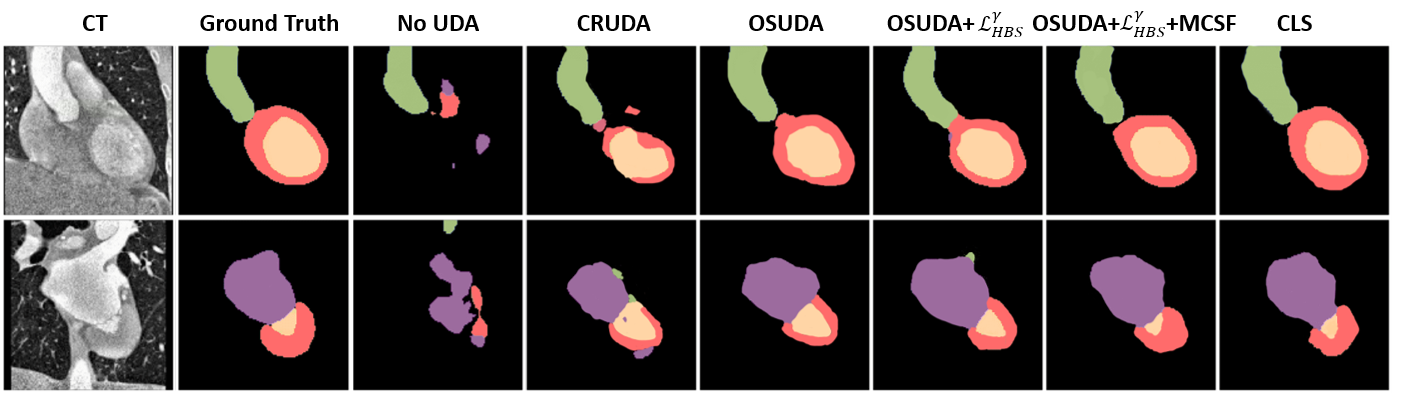}%
\end{center} 
\caption{Comparisons of our framework against other UDA methods, and ablation studies for cardiac MR to CT segmentation. The source-available CLS \citep{liu2021adversarial} is regarded as an ``upper bound."} 
\label{cardfig}
\end{figure*}

\subsubsection{Cross-subtype UDA: HGG to LGG}
The subtypes of HGG and LGG can take different tumor sizes and positions. Following the standard cross-subtype data split protocol in \cite{shanis2019intramodality} to adapt the model trained on HGG to LGG subjects. We chose the same backbone as in \cite{shanis2019intramodality} with 15 layers in U-Net \citep{ronneberger2015u}. Networks were trained with four-channel sliced 2D axial MRI slices to perform pixel-wise four-class segmentation, including background, EnhT, CoreT, and ED.  The input samples have the size of 128$\times$128$\times$4, which is a slice-wise concatenation of four MRI modalities. The networks were trained with a batch size of 12. The consecutive iteration in the memory $H$ was set as 3. Following \cite{shanis2019intramodality}, both HGG and LGG volumes were split into training and testing datasets. We pre-trained our network on the source domain data over 150 epochs as in \cite{shanis2019intramodality}. The adaptation training took about 2 hours.

In Table~\ref{tab1}, we provide the results of our quantitative evaluations. Due to different class-wise pixel proportions in the two subtypes, the class-ratio-based CRUDA \citep{bateson2020source} only achieved marginal improvements. Especially, the DSC of CoreT was inferior to the source model without adaptation, which can be the case of the negative transfer \citep{wang2019characterizing}. Our proposed OSUDA with $\mathcal{L}_{HBS}^{\gamma}$ and MCSF was able to achieve superior performance for source-free UDA segmentation, and approached the results of source-available SEAT~\citep{shanis2019intramodality}. The ablation studies also confirmed the effectiveness of the adaptive channel-wise weighting, SE minimization, $\mathcal{L}_{HBS}^{\gamma}$, and MCSF. 

The qualitative comparisons are shown in Fig. \ref{exp1}. Source-free UDA methods were able to substantially improve the performance over the source domain model without adaptation. However, CRUDA \citep{bateson2020source} tended to predict a larger area for the tumor.


\subsection{Segmentation of Cardiac Structures}

We used the MM-WHS database for the whole heart segmentation  \citep{zhuang2016multi}, which contains a total of 40 datasets acquired from multiple clinical sites. There are a total of 20 subjects who have MRI scans, each of which has a total of 128 MRI slices. Another 20 subjects have CT scans, each of which has about a total of 256 slices. The segmentation label of each slice in the cardiac datasets has five classes, including left ventricle blood cavity (LVC), left atrium blood cavity (LAC), the myocardium of the left ventricle (MYO), ascending aorta (AA), and background. Note that the subjects with MRI and CT scans are unpaired. We followed the previous evaluation protocols in \cite{chen2019synergistic,zou2020unsupervised,chanti2021optimal} to use the coronal view slices, and chose 8/2 subject split for training and testing, respectively. We chose MRI and CT as the source and target domains, respectively. As a preprocessing step, the slice was cropped to $256\times256$. For a fair comparison, we adopted the same backbone in \cite{chen2019synergistic,zou2020unsupervised}. We set $\varphi=5$ and linearly decreased $\lambda$ from 10 to 0. In all of our cardiac segmentation tasks, we set the batch size $B$ as 12, and the consecutive iteration in the memory $H$ as 5. We trained the source model over 100 epochs, followed by the source-free adaptation over 100 epochs. The training took about 1.5 hours.

The numerical comparisons are provided in Table \ref{card1}. Consistent with the brain tumor segmentation tasks, our framework performed better than CRUDA \citep{bateson2020source} and SFKT \citep{liu2021source}, by more than 4\% w.r.t. DSC, and 6mm w.r.t. HD. When compared with the methods with the source data, our framework performed better than the classical cycleGAN \citep{zhu2017unpaired}.  In Fig. \ref{cardfig}, we provide the qualitative comparisons. 

\begin{table}[t!]
\caption{Comparisons of cardiac MR to CT segmentation with only one target domain subject for training as in OLVA \citep{chanti2021optimal}. $\pm$ indicates standard deviation. The UDA methods with source domain data are regarded as ``upper bounds."} 
\label{card2}
\centering
\resizebox{1\linewidth}{!}{
\begin{tabular}{l|c|ccccc}
    \hline
    \multirow{2}*{Method}&Source& \multicolumn{5}{c}{DSC [\%] $\uparrow$} \\ \cline{3-7}
    &data& AA & LAC & LVC & MYO & Average   \\ \hline \hline
     
    CRUDA  & Partial & 53.45 &59.20& 73.32& 35.64&55.40$\pm$0.13  \\
    SFKT  & \textbf{no} & 51.38 &56.93& 71.30&33.76&53.34$\pm$0.14 \\  \hline

    \textbf{OSUDA} & \textbf{no} &{57.96} & {63.37}& {77.85}&  39.82  &{59.75$\pm$0.11} \\ 
    
    {-AC} & \textbf{no} &57.63 & 62.03& 77.45&  39.74  &59.21$\pm$0.15 \\
    
    {-SE} & \textbf{no} &57.13 & 62.41& 77.32&  39.26  &59.03$\pm$0.13 \\

     +$\mathcal{L}_{HBS}^{\gamma}$ & \textbf{no} &{58.06} & {63.49}& {78.12}& 39.94&{59.90$\pm$0.12}\\     
     
     +$\mathcal{L}_{HBS}^{\gamma}$+MCSF & \textbf{no} &\textbf{58.27} & \textbf{64.18}& \textbf{78.44} &  \textbf{40.51} &\textbf{60.35$\pm$0.08} \\    \hline   \hline

SIFA & Yes& 62 &53 &80 &39 &62 \\

OLVA & Yes& 60 & 70  & 78  & 68 & 69\\

\hline 

\end{tabular}
}
\end{table}

In addition, we followed the evaluation protocol in \cite{chanti2021optimal} to evaluate the performance with only one subject in the target domain. As in \cite{chanti2021optimal}, we took three adjacent slices ($256 \times 256 \times 3$) as input and predicted the segmentation mask of the middle slice to explore the 2.5D information. The segmentation network has 5 convolutional and 5 de-convolutional layers \citep{chanti2021optimal}. We set $H=3$ and $\varphi=5$, and linearly decreased $\lambda$ from 10 to 0.

The numerical comparisons are provided in Table \ref{card2}. We can see that the performance improvements over CRUDA \citep{bateson2020source} and SFKT \citep{liu2021source} are more appealing. Specifically, the average DSC of MCOSUDA was about 15\% higher than CRUDA and SFKT.

\begin{figure}[t]
\begin{center} \vspace{+10pt}
\includegraphics[width=1\linewidth]{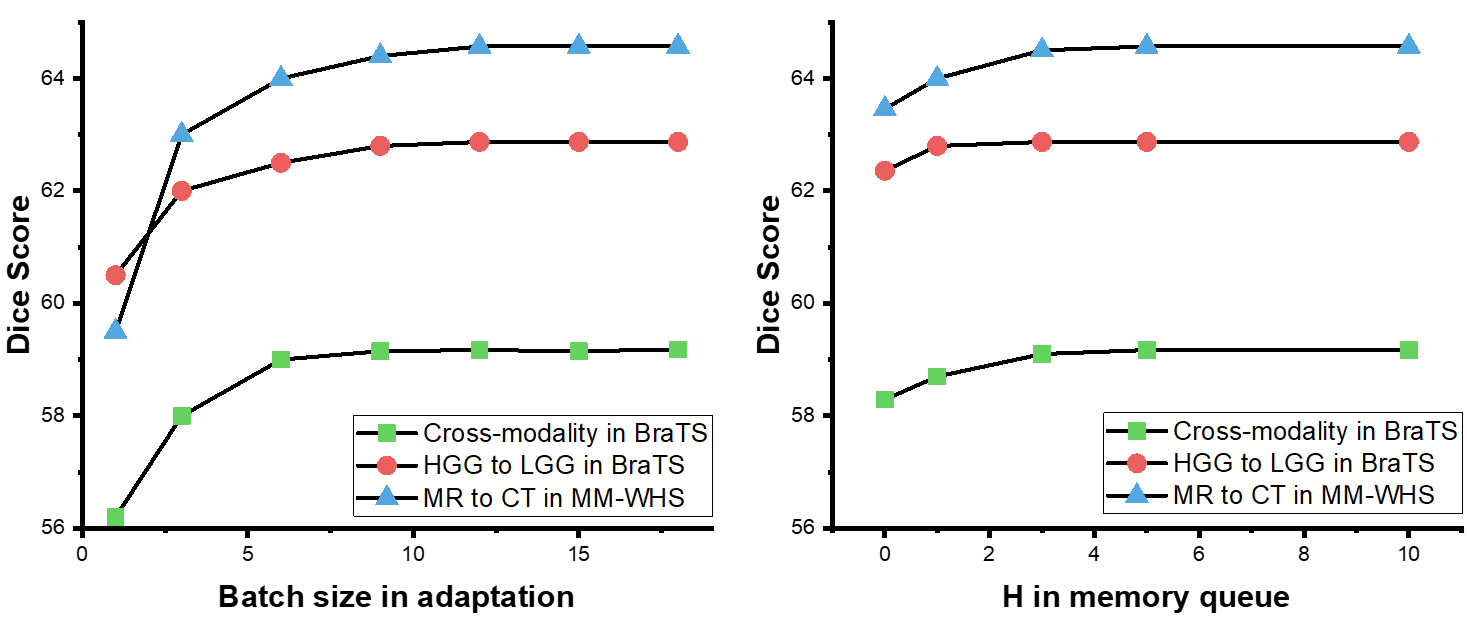}
\end{center}  
\caption{The average DSC of OSUDA+$\mathcal{L}_{HBS}^{\gamma}$+MCSF (MCOSUDA) with different batch size (left) and $H$ in the memory queue for the cross-modality (blue) and the HGG to LGG (red) brain tumor segmentation task, and the cardiac MR to CT segmentation task (green).} 
\label{sens1}\end{figure}

\subsection{Sensitive analysis of hyper-parameters}  

There are several hyper-parameters involved in our framework. In this subsection, we provide a systematical analysis of these hyper-parameters. Firstly, the stability of the BN statistics can be highly associated with the batch size. A larger batch size can provide more unbiased statistics, while introducing more computation and memory costs. As shown in Fig. \ref{sens1} left, the performance can be benefited, by increasing the batch size from 1 to 10, while the performance is almost stable for the batch size larger than 10 for all of the tasks. Therefore, we simply set all of the batch size to 12 in this work.

\begin{table}[t] 
\caption{The average DSC of OSUDA with fixed or linear changed $\lambda$ for the BraTS cross modality/subtype and MR to CT segmentation tasks.} \vspace{+5pt}
\label{senstab1}
\centering
\resizebox{0.85\linewidth}{!}{
\begin{tabular}{c|c|c|c|c|c|c}
    \hline
     \multicolumn{7}{c}{Cross modality segmentation task using BraTS18} \\\hline
     $\lambda$ & 10$\rightarrow$0  & 10 & 5 & 3  & 1 & 0\\\hline
     DSC & \textbf{57.84}   & 56.98  & 57.15  & 57.26  & 57.04   &  {56.58} \\\hline  \hline   
     
     \multicolumn{7}{c}{HGG to LGG segmentation task using BraTS18} \\\hline
     $\lambda$ & 10$\rightarrow$0  & 10 & 5 & 3  & 1 & 0\\\hline
     DSC & \textbf{61.94}   & 60.97  & 61.02  & 61.14  & 61.08   & 60.78 \\\hline\hline

     \multicolumn{7}{c}{MR to CT segmentation task using MM-WHS} \\\hline
     $\lambda$ & 10$\rightarrow$0  & 10 & 5 & 3  & 1 & 0\\\hline
     DSC & \textbf{63.11}   & 62.95  & 62.96  & 62.95  & 62.90   & 62.87 \\\hline
         
\end{tabular}
} 
\end{table}

\begin{figure}[t]
\begin{center} 
\includegraphics[width=1\linewidth]{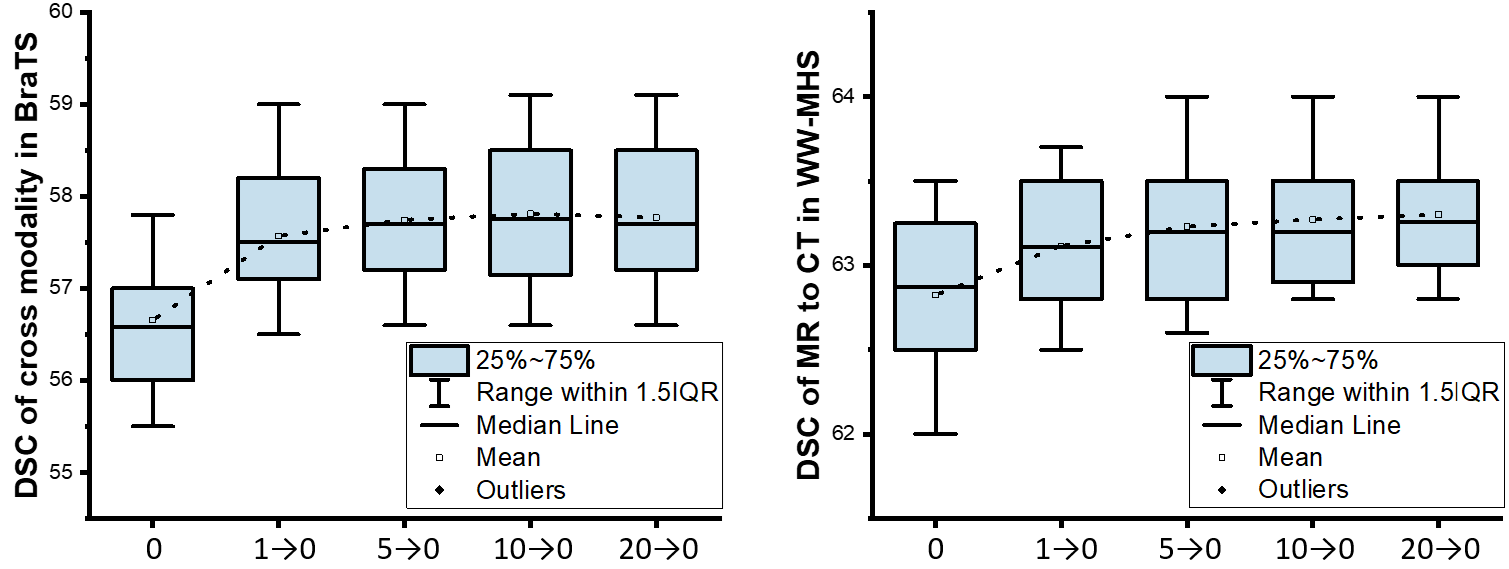}
\end{center}  
\caption{The DSC of OSUDA+$\mathcal{L}_{HBS}^{\gamma}$+MCSF (MCOSUDA) in cross-modality tumor segmentation (left) and cardiac MR to CT segmentation (right) with different $\lambda$.} 
\label{sens2}\end{figure}

We used $\lambda$ to weight the SE minimization, which is linearly changed 
from a relatively large positive value to 0 at the adaptation stage. As shown in Table~\ref{senstab1}, we can see that the linearly decreasing $\lambda$ can be better than fixed $\lambda$. To select the proper initial value of $\lambda$, we provide the sensitivity analysis in Fig. \ref{sens2}. The range of linearly decreasing $\lambda$ can be relatively stable from 5$\rightarrow$0 to 15$\rightarrow$0. We simply set  $\lambda$ to 10$\rightarrow$0 for all of our experiments. We note that setting $\lambda$ to 0 is equivalent to OSUDA-SE, i.e., without the SE minimization term.

In MCSF, we utilized a memory queue to measure the historical consistency. As shown in Fig. \ref{sens1} right, we analyzed the relationship between $H$ and the performance. In both the cross-modality segmentation and the cardiac MR to CT segmentation tasks, the performances were stable, since $H$ had a value larger than 5. For the HGG to LGG task, $H=3$ was sufficient to achieve the DSC of 62.87\%.

\begin{figure}[t]
\begin{center} 
\includegraphics[width=1\linewidth]{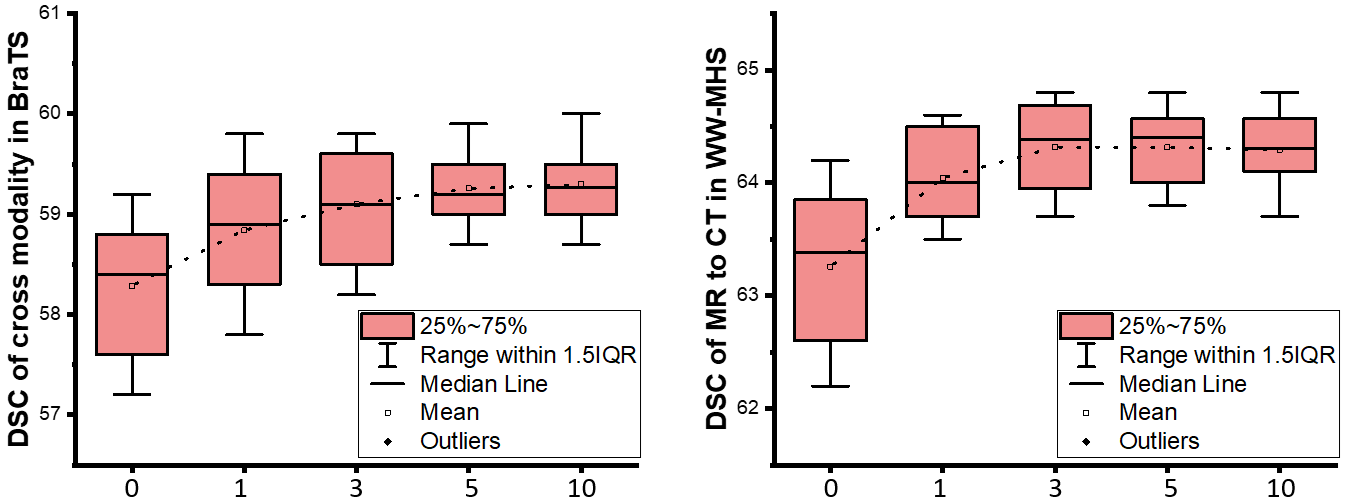}
\end{center}  
\caption{The DSC of OSUDA+$\mathcal{L}_{HBS}^{\gamma}$+MCSF (MCOSUDA) in cross-modality tumor segmentation (left) and cardiac MR to CT segmentation (right) with different $\varphi$.} 
\label{sens3}\end{figure}

In order to balance the MCSF loss, we used $\varphi$ to weight the objectives. For the sensitivity analysis, we provide the comparisons with different $\varphi$ as shown in Fig. \ref{sens3}. By setting $\varphi$ larger than 5, we were able to achieve relatively stable performance for all of the tasks, e.g., the average DSC of 59.17\% in cross-modality whole tumor segmentation using the BraTS2018 dataset. 

{In addition, $\lambda ^n$ is set to select the top $\alpha\%$ of the most confident pixels in each class. Therefore, $\alpha$ is a to-be tuned hyperparameter, and $\lambda^n$ adaptively changes in each iteration based on $\alpha$. Increasing $\alpha$ along with the training is a typical solution to accommodate the noisy level change of pseudo-label \citep{zou2019confidence,liu2020energy,liu2021generative}. In Table \ref{senstab2}, we compared the model with different start and end $\alpha$ for both tasks. We can see that too large end $\alpha$ can lead to worse performance. Since the pseudo-label can still be noisy at the late training epochs, assuming more than 80\% of them are correct can mislead the training. In addition, too small start $\alpha$ can lead to slower convergence, since the pseudo label is not sufficiently utilized.}

\section{Discussion}

The problem of domain shift is prevalent, when applying deep learning models trained on source domain data to carry out a variety of tasks in the target domain. As a result, the performance degradation has been clearly observed in tasks using data from different centers \citep{liu2021subtype}, scanners \citep{ghafoorian2017transfer}, populations, subtypes \citep{liu2021subtype}, and modalities \citep{liu2021generative}. Among these cases, cross-modality UDA can be the most challenging task, due to different imaging principles involved, resulting in different image appearances. The general validity and efficacy of our proposed framework were demonstrated using cross-MR-modality and cross-subtype brain tumor segmentation tasks, and cardiac MR to CT segmentation task. While these databases presented different challenges, our framework robustly achieved superior performance consistently compared with the conventional approaches. In addition, different backbones were adopted, similar to the previous works. Notably, our proposed framework is not dependent on prior knowledge of the specific imaging modality, and can be easily applied to other UDA tasks.

Due to concerns over patient data privacy and IP, the restriction of the source data sharing in clinical practice can be a significant obstacle for many UDA approaches. To address the issue, in this work, we proposed a novel UDA segmentation framework in the absence of source domain samples. To the best of our knowledge, this is one of the first attempts at source-relaxed UDA for image segmentation, which does not need an auxiliary network, or the unreliable assumption of the same label proportion \citep{bateson2020source}. Our framework is only reliant on a pre-trained OS segmentor, with the widely used BN.

As shown in Tables 2-4 and Figs. 6-8, our framework yields superior performance over the other comparison methods, when qualitatively and quantitatively evaluated. There are a few important differences that give insights into their performances. We can see that the performance degradation of the source domain model widely exists, especially when we apply our framework to a different target domain. Due to a relatively large domain shift in the cardiac MR to CT segmentation task, the DSC of the source domain model was only 17.2\%. Since the multi-modality segmentation task is widely used in clinical practice, and the dual annotation on multiple modalities can be a large burden, the domain adaptation methods can be a viable solution \citep{chen2019synergistic}.

CRUDA \citep{bateson2020source} assumes that the pixel proportion is consistent between two domains. In adaptation, this prior knowledge was used as the only transferable information. In Table \ref{tab1}, we show that the performance of CRUDA is largely inferior to our framework, since the tumor pixel proportion is largely different between HGG and LGG subjects. In addition, the class-ratio prediction model is usually not ``off-the-shelf," since it is not typically used in clinical practice and needs specialized training in the source domain. Recently, there are several contemporary or follow-up works, aimed at source-free UDA for natural image segmentation, which are re-implemented and compared in Tables \ref{tab2}, \ref{tab1}, and \ref{card1}. Although the relatively good performance was achieved by \cite{kundu2021generalize}, its domain generalization training in the source domain limits the model to be used for source-free UDA. The additional attention network used in SFKT may require relatively more data for training, thus yielding additional parameters. For the cardiac MR to CT segmentation task, with only one target training sample, the DSC of SFKT is 7\% lower than our framework. The denoising strategy used in \cite{chen2021source} utilized the Monte Carlo method, which is challenging for Bayesian uncertainty approximation and training \citep{liu2021generative}. 

Several state-of-the-art source-available UDA segmentation works are also compared \citep{liu2021adversarial,zou2020unsupervised}, which are used as the ``upper bounds." Note that we did not manage to outperform all of them, since we have different settings when training on the source data, and it can be applied to different applications. Nonetheless, the fact that our framework approached their performance indicates the superior performance of our source-free UDA. In some of the tasks, our framework was able to outperform even the source data available methods \citep{zhu2017unpaired,shanis2019intramodality,chen2019synergistic}, which further demonstrates the superiority of our framework.

\begin{table}[t] 
\caption{{The average DSC of MCOSUDA with different $\alpha$ for three tasks.}}  
\label{senstab2}
\centering
\resizebox{0.85\linewidth}{!}{
\begin{tabular}{c|c|c|c|c|c}
    \hline
     \multicolumn{6}{c}{{Cross modality segmentation task using BraTS18}} \\\hline
     $\alpha$ & 20$\rightarrow$80  & 10$\rightarrow$80 & 30$\rightarrow$80 & 20$\rightarrow$70  & 20$\rightarrow$90 \\\hline
     DSC & \textbf{59.17}   & 59.02  & 58.87  & 59.13  & 58.74    \\\hline  \hline   
     
     \multicolumn{6}{c}{{HGG to LGG segmentation task using BraTS18}} \\\hline
     $\alpha$ & 20$\rightarrow$80  & 10$\rightarrow$80 & 30$\rightarrow$80 & 20$\rightarrow$70  & 20$\rightarrow$90 \\\hline
     DSC & \textbf{62.87}   & 62.74  & 62.65  & 62.72  & 62.46    \\\hline\hline
     
     \multicolumn{6}{c}{{MR to CT segmentation task using MM-WHS}} \\\hline
     $\alpha$ & 20$\rightarrow$80  & 10$\rightarrow$80 & 30$\rightarrow$80 & 20$\rightarrow$70  & 20$\rightarrow$90 \\\hline
     DSC & \textbf{64.57}   & 64.50  & 64.48  & 64.49  & 64.42    \\\hline
        
\end{tabular}
} 
\end{table}

\begin{figure}[t]
\begin{center} 
\includegraphics[width=1\linewidth]{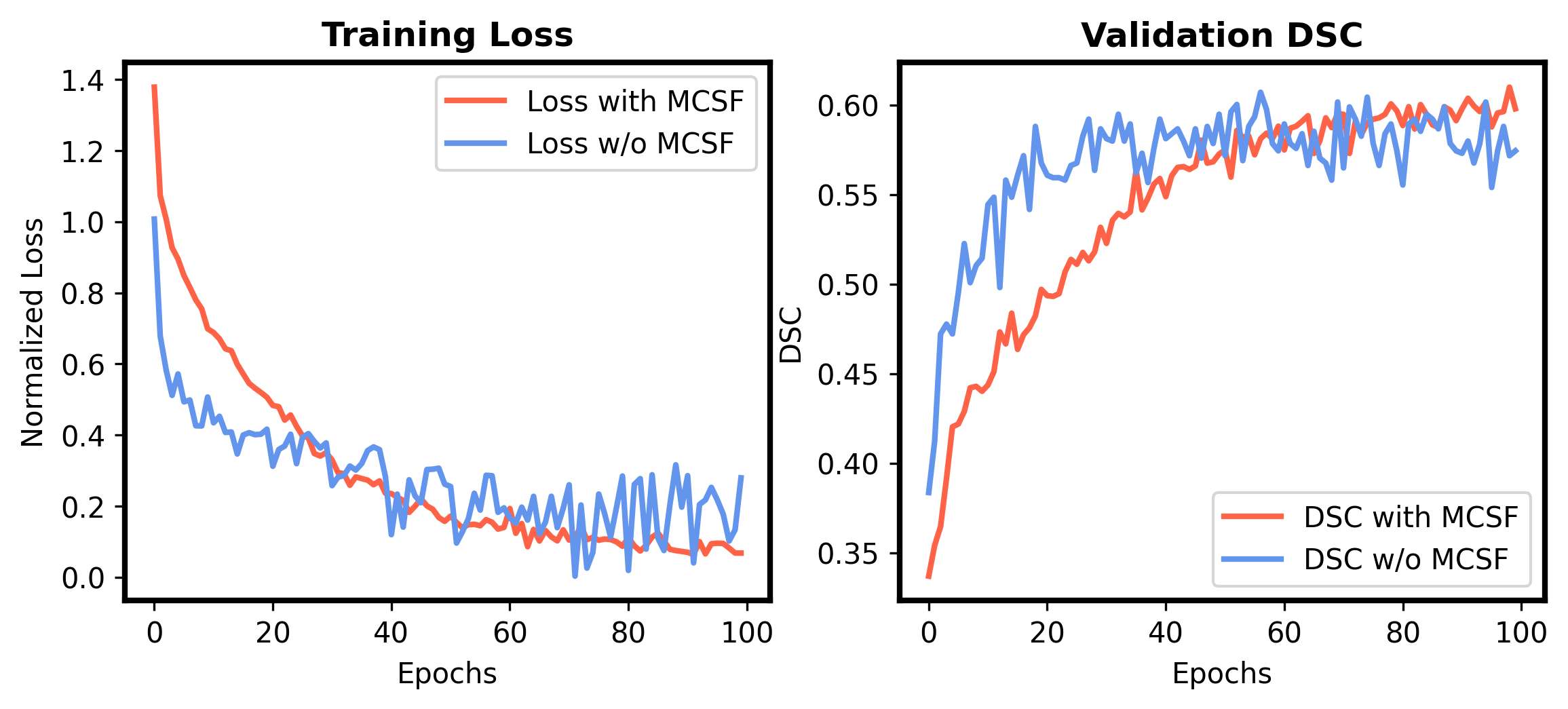}
\end{center}   
\caption{The training loss (left) and validation DSC (right) of our OSUDA, with or without MCSF in cross-MR-modality tumor segmentation.}
\label{variation}\end{figure}

We proposed to explore the BN statistics for source-free UDA using a systematical framework with low-order statistics progression and high-order statistics consistency loss. The transferability of each channel was exploited for the channel-wise adaptive alignment. We further proposed to utilize both the low-order statistics discrepancy and scaling factor for the transferability quantification, which are evidenced by the ablation studies for OSUDA-AC and OSUDA+$\mathcal{L}_{HBS}^{\gamma}$, respectively. In addition to the BN statistics alignment, we further integrated the SE minimization and MCSF into a unified framework to exploit the unsupervised learning and memory-based learning, respectively. The unsupervised SE minimization and pseudo label used in self-training can be noisy and unstable \citep{zou2019confidence,liu2021generative,liu2020energy}. In our source-free setting, no source domain data are available at the adaptation stage to correct the biased update similar to the source available self-training \citep{zou2019confidence}, which makes the task more challenging. To address this issue, we proposed a novel MCSF strategy to efficiently regularize the adaptation with memory-based learning. In Fig. \ref{variation}, we show that MCSF was able to effectively stabilize the training, and achieve better performance.   

{This work aimed to develop a source-free UDA approach, in which source domain data are not available at the adaptation stage, which is considered a more restricted setting, compared with UDA with source data. This setting, in turn, could help avoid any potential issues over patient data sharing or IP. As such, the evaluations against UDA approaches, in which source data are available, are used for performance comparison purposes only. }

There are a few important aspects that are not fully studied in the present work. First, we proposed to exploit the BN statistics. Recent studies \citep{zhou2021domain} used instance-wise statistics for domain generalization, which can be orthogonal to our framework and could be potentially added to our framework. Second, we only investigate the case where the source and target domain models use the same backbone, which is a typical setting in UDA. The different backbones may be achievable by initializing the target model with knowledge distillation \citep{hinton2015distilling}. Third, we have taken the correlation between neighboring MRI slices into consideration (i.e., 2.5D segmentation) as in \cite{chanti2021optimal} in Table \ref{card2}. Yet, the 3D segmentation backbones can hardly be applied to the datasets, due to the relatively limited number of 3D datasets to provide reliable BN statistics. In addition, while our experiments showed that many of the weights were not sensitive, they needed to be carefully tuned to balance among the optimization objectives.

\section{Conclusion} 
 
In this paper, we have proposed a novel and practical source-free UDA framework, aimed at image segmentation. Our framework was only reliant on an OS pre-trained segmentation network with BN in the source domain, which could thereby sidestep the concerns over the patient data sharing and IP inherent in conventional UDA. The BN statistics were systematically investigated for domain alignment. Specifically, the low-order BN statistics progression with EMD was proposed to gradually learn the target domain-specific mean and variance. The domain shareable high-order BN statistics consistency was encouraged by the HBS loss, which was adaptively adjusted, according to the channel-wise transferability. In addition to quantifying the transferability based on low-order statistics discrepancy, the high-order scaling factor was further explored. An unsupervised learning objective, i.e., SE minimization, was incorporated into our framework, and the novel queued memory-consistent self-training was further proposed to achieve stable memory learning with the pseudo label. Extensive experiments, ablation studies as well as sensitivity analysis on the cross-MR-modality and cross-subtype brain tumor segmentation tasks and cardiac MR to CT segmentation task demonstrated the effectiveness of our OSUDA, which can be potentially applied in a clinically meaningful setting.

\section*{Acknowledgments}

We gratefully acknowledge funding support from NIH R01DC018511, R01DE027989, and P41EB022544.

\bibliographystyle{model2-names.bst}\biboptions{authoryear}
\bibliography{refs}

\end{document}